\documentclass[letterpaper]{article} 

\usepackage[preprint]{aaai2027}

\usepackage[hyphens]{url}
\usepackage{graphicx}

\usepackage{natbib}
\usepackage{caption}

\usepackage{amsmath,amssymb,amsfonts}
\usepackage{booktabs}
\usepackage[table]{xcolor}

\definecolor{tablegray}{gray}{0.93}

\title{QCPruner: Query-Conditioned Population Coverage for Visual Token Pruning}

\author{
Shengli He\textsuperscript{\rm 1},
Yongchao Liang\textsuperscript{\rm 1}\corresponding,
Roumeng He\textsuperscript{\rm 2},\\
Junjie Zeng\textsuperscript{\rm 1},
Jiyuan He\textsuperscript{\rm 1},
Can Wu\textsuperscript{\rm 1},
Li Zheng\textsuperscript{\rm 1}
}

\affiliations{
\textsuperscript{\rm 1}Guizhou University, Guiyang, China\\
\textsuperscript{\rm 2}Shanghai Ocean University, Shanghai, China\\
Shengli He: shengli5930@gmail.com,
Yongchao Liang: ycliang@gzu.edu.cn
}

\begin{document}

\maketitle

\begin{abstract}
The high visual-token load in multimodal large language models (MLLMs) motivates training-free pruning to reduce later-layer computation, but under a fixed budget, pruning must preserve query-relevant evidence while avoiding redundancy.
Existing methods rank tokens, diversify selected subsets, or optimize coverage without using a shared per-visual query utility to weight both visual targets and candidate representatives.
We introduce \textbf{QCPruner}, which makes both roles query-conditioned through bilateral utility weighting.
Using keyword-matched query anchors, QCPruner fuses two cross-modal cues into utility and applies it to both visual targets and candidate representatives within visual-affinity-based coverage.
The resulting nonnegative facility-location objective is monotone and submodular, retains the standard $(1-1/e)$ greedy guarantee, and requires no model training or parameter updates.
Across LLaVA-1.5, LLaVA-NeXT, LLaVA-Video, and Qwen2.5-VL, QCPruner achieves the highest average relative performance among evaluated complete-system pruning methods at every reported token budget.
At 32 of 576 tokens on LLaVA-1.5-7B, it retains 96.1\% of unpruned performance, versus 93.9\% for the strongest evaluated baseline. At 256 of 1296 tokens on Qwen2.5-VL-7B, the corresponding values are 96.7\% and 92.5\%.
\end{abstract}

\section{Introduction}

MLLMs encode images or videos as visual-token prefixes and process them with a user instruction in a Transformer language model~\cite{attention,llava,llava15,llava_next,qwen2vl,qwen25vl}.
A standard LLaVA-1.5 image yields 576 visual tokens, while high-resolution and video inputs generate thousands.
These prefixes increase prefill computation and activation memory, motivating training-free pruning.
Pre-decoder methods avoid processing discarded tokens throughout the decoder but rely on signals formed before decoder-side contextual cross-modal reasoning~\cite{visionzip,divprune,mmtok,prunesid}.
In-decoder methods obtain contextual cross-modal states only after preceding layers process the full sequence~\cite{fastv,sparsevlm}.
Regardless of where pruning is applied, the practical goal is to preserve task performance at a fixed final token budget, rather than merely removing more tokens~\cite{right_problem}.

Independent attention- or relevance-based ranking can retain redundant neighboring patches~\cite{fastv,sparsevlm}.
Redundancy- and diversity-oriented methods preserve visual breadth, but visual redundancy alone need not identify evidence required by the current instruction~\cite{visionzip,dart,divprune}.
Conditional DPPs couple instruction relevance with selected-set diversity, yet focus on the retained subset and its internal interactions rather than explicit representation of the broader visual population~\cite{cdpruner}.
Coverage objectives measure representation against a broader target population.
MMTok instantiates this approach with coverage over text and visual target populations~\cite{mmtok}. 
It is query-aware and population-relative, but introduces the query as a separate text-side target rather than directly conditioning visual-to-visual coverage. 
We therefore ask: how should contextual query evidence determine what deserves coverage and what can serve as a representative within the visual population itself?

\textbf{QCPruner} addresses this with a target--representative formulation.
Each visual target is weighted by its query utility, and each retained representative is likewise weighted by its query utility, with visual affinity linking the two roles. 
Target utility emphasizes query-relevant evidence to preserve, while representative utility favors query-relevant tokens for covering similar visual content.
QCPruner uses contextual per-visual query utility to weight both sides of visual-to-visual coverage, whereas MMTok sums text--vision and vision--vision facility-location terms, and conditional-DPP methods model interactions within the selected subset.
Specifically, QCPruner derives contextual utility from keyword-matched query anchors and complementary cross-modal cues, then performs coverage-aware greedy selection under a fixed token budget.

Across four backbones spanning standard images, high-resolution images, and video, QCPruner achieves the highest average relative performance among evaluated complete systems across all reported budgets. 
It retains 96.1\% at 32/576 LLaVA-1.5 tokens, 96.7\% at 256/1296 Qwen2.5-VL tokens, and 96.3\% at 1024/10816 LLaVA-Video tokens. 
Matched-budget comparisons further give higher Avg. Rel. than MMTok across all three image backbones.
We report retention--cost trade-offs rather than claiming uniform speed superiority.

Our contributions are summarized as follows:
\begin{itemize}
\item We introduce bilateral contextual query weighting into coverage-based selection: query utility weights both visual targets and candidate representatives, while visual affinity links the two roles. The resulting nonnegative objective preserves the standard greedy guarantee.
\item Without training the MLLM or a pruning module, we derive contextual utility from keyword-matched prompt positions by combining hidden-state similarity and pre-softmax query--key alignment.
\item We validate QCPruner across four MLLM backbones and twelve budget settings, covering standard images, high-resolution inputs, and video. Controlled visual-only, one-sided, and selector comparisons separately examine coverage weighting, bilateral utility, and subset selection.
\end{itemize}

\section{Related Work}

\paragraph{MLLM visual-token layouts.}
MLLMs project features from visual encoders such as CLIP or SigLIP into a language model and process them jointly with textual instructions~\cite{clip,siglip,llava,llava15,minigpt4,internvl,mllm_survey}.
Their visual-token layouts range from fixed image grids to high-resolution tiles, repeated video-frame grids, and model-specific arrangements~\cite{llava_next,qwen25vl}.
These layouts differ in token count, spatial organization, and structural boundaries, leading to different token scales and structural constraints for visual-token pruning.

\paragraph{Training-free visual-token pruning.}
Training-free methods remove or merge visual tokens using attention, saliency, similarity, redundancy, or cross-modal relevance~\cite{fastv,sparsevlm,prumerge,visionzip,similarity_pruning,fitprune,dart,prunesid}.
Pre-decoder methods reduce the sequence before language-model processing, while in-decoder methods exploit contextual cross-modal states formed by preceding layers.
Many ranking-based approaches score tokens independently, which can retain individually salient but mutually redundant visual evidence.
This motivates subset-level objectives that explicitly consider relations among retained tokens.

\paragraph{Subset-level visual-token selection.}
Subset-level methods model interactions among selected tokens or how well they represent a broader visual population.
DivPrune optimizes visual diversity, while CDPruner combines instruction relevance with conditional-DPP selection to diversify the retained subset~\cite{divprune,cdpruner}.
MMTok instead combines text--vision and vision--vision facility-location coverage, making selection both query-aware and population-relative~\cite{mmtok}.
QCPruner builds on population-level coverage but differs by deriving contextual per-visual query utility at an in-decoder hook and applying it to both target importance and representative suitability within a single visual population, thereby directly query-conditioning visual-to-visual coverage.

\section{Method}
\label{sec:method}

\begin{figure*}[t]
    \centering
    \includegraphics[width=\textwidth]{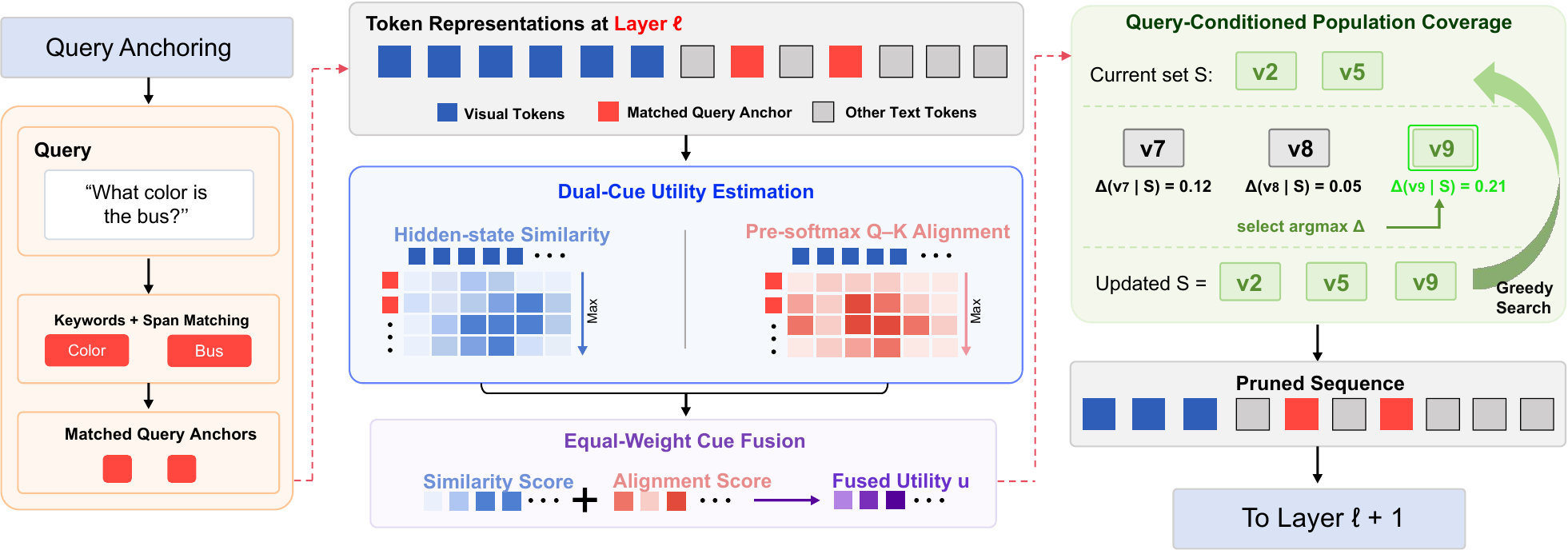}
    \caption{
    Overall framework of QCPruner.
    At the post-layer-$\ell$ hook, matched query anchors are used to compute hidden-state cosine similarity from the layer output and pre-softmax query--key alignment recorded within the same layer.
    Their fused utility weights both the target and representative sides of visual affinity, and greedy marginal gain selects $K$ representatives to maximize coverage of the full candidate population.
    Structural tokens excluded from the candidate set are preserved. The figure uses $S$ for the current partial set and $\mathcal{K}$ for its final counterpart in the text.
    }
    \label{fig:overall_framework}
\end{figure*}

We first formulate visual-token pruning under a fixed token budget in Section~\ref{sec:problem_setup}. 
Section~\ref{sec:query_anchor} then constructs keyword-matched query anchors to identify informative prompt positions, 
and Section~\ref{sec:alignment_enhanced_scoring} estimates contextual visual-token utility by combining hidden-state similarity with query--key alignment. 
Finally, Section~\ref{sec:coverage_selection} introduces query-conditioned population coverage, 
where the resulting utility weights both target and representative roles and greedy marginal-gain selection produces the retained subset. 
Figure~\ref{fig:overall_framework} illustrates the overall framework of QCPruner. The entire procedure is training-free and requires no model-parameter updates.

\subsection{Problem Setup}
\label{sec:problem_setup}

Let $\mathcal{V}=(v_1,\ldots,v_{N_v})$ be the ordered candidate patch tokens and $\mathcal{T}=(t_1,\ldots,t_{L_t})$ the prompt tokens.
At the post-layer-$\ell$ hook, where $\ell$ denotes the zero-based decoder-layer index, contextual states $\mathbf{h}^{v,\ell}_i,\mathbf{h}^{t,\ell}_j\in\mathbb{R}^{d}$ are output by layer $\ell$ and pruning precedes layer $\ell+1$.
Model-specific separators and other non-patch structural tokens are excluded from $\mathcal{V}$ and remain in the sequence.
We select indices
\begin{equation}
\mathcal{K}\subseteq \{1,\ldots,N_v\}, \qquad |\mathcal{K}|=K,
\label{eq:budget_constraint}
\end{equation}
where $K<N_v$ is the patch-token budget.

Every $v_i$ has two possible roles: as a \emph{target}, its utility measures how much preserving its evidence matters, while as a \emph{representative}, its utility and affinity determine how credibly it can represent other targets when selected.
Sections~\ref{sec:query_anchor}--\ref{sec:alignment_enhanced_scoring} estimate these query-conditioned utilities, and Section~\ref{sec:coverage_selection} optimizes representation of the full weighted population.

\subsection{Keyword-Matched Query Anchors}
\label{sec:query_anchor}

System scaffolding and generic instruction text can dilute conditioning aggregated over the full prompt.
From the user-visible prompt, KeyBERT~\cite{keybert} therefore extracts at most six ranked content unigrams $\Phi=\{\phi_m\}_{m=1}^{M}$.
These strings only locate prompt positions: we retokenize each keyword with the MLLM tokenizer, match it as a contiguous subsequence of the original prompt, and retain all matched occurrences.

Let $\mathbf{z}=(z_1,\ldots,z_{L_t})$ be the prompt token-ID sequence, and let $\mathcal{S}=\{(a_s,b_s)\}_{s=1}^{R}$ collect the inclusive spans of all successful keyword occurrences, including repeated occurrences of the same keyword.
The anchor set contains positions rather than token IDs:
\begin{equation}
\mathcal{I}_{\mathrm{anc}}=
\bigcup_{s=1}^{R}
\{a_s,a_s+1,\ldots,b_s\}.
\label{eq:anchor_set}
\end{equation}
This position-based definition handles keywords split into multiple MLLM subwords.
If no keyword matches, we deterministically fall back to user-content positions after excluding special tokens, visual placeholders, and known template spans.
The matched positions index contextualized MLLM states for subsequent scoring.

\subsection{Dual-Cue Query-Conditioned Utility Estimation}
\label{sec:alignment_enhanced_scoring}

At the post-layer-$\ell$ hook, the semantic cue uses contextual states output by decoder layer $\ell$, whereas the alignment cue uses the pre-softmax query and key projections recorded while executing that same layer.
We omit the layer index $\ell$ from subsequent notation.
For each $v_i$, the following two scores are proxies for compatibility with at least one anchor in $\mathcal{I}_{\mathrm{anc}}$.

\paragraph{Similarity-based semantic cue.}
The similarity cue uses cosine proximity in the contextual hidden space as a representation-level proxy.
We first normalize the hidden states:
\begin{equation}
\bar{\mathbf{h}}^{t}_j=
\frac{\mathbf{h}^{t}_j}{\|\mathbf{h}^{t}_j\|_2},
\qquad
\bar{\mathbf{h}}^{v}_i=
\frac{\mathbf{h}^{v}_i}{\|\mathbf{h}^{v}_i\|_2}.
\end{equation}
The cosine similarity between anchor $t_j$ and visual token $v_i$ is
\begin{equation}
s_{j,i}^{\mathrm{sim}}
=
(\bar{\mathbf{h}}^{t}_j)^{\top}\bar{\mathbf{h}}^{v}_i.
\end{equation}
We define the similarity-based relevance of $v_i$ as
\begin{equation}
r_i^{\mathrm{sim}}
=
\max_{j\in\mathcal{I}_{\mathrm{anc}}}
s_{j,i}^{\mathrm{sim}}.
\label{eq:method_sim}
\end{equation}
Max aggregation retains a high score when any matched anchor aligns with a token and avoids diluting strong anchor-specific relevance through averaging.

\paragraph{Query--key alignment cue.}
Let $\mathbf{q}^{(h)}_j$ and $\mathbf{k}^{(h)}_i$ be the query and key vectors for anchor $t_j$ and visual token $v_i$ at head $h$, where $H$ is the number of heads and $d_h$ is the dimensionality of each head.
We use the head-averaged, scaled dot product before softmax:
\begin{equation}
A_{j,i}
=
\frac{1}{H}
\sum_{h=1}^{H}
\frac{
(\mathbf{q}^{(h)}_j)^\top \mathbf{k}^{(h)}_i
}{
\sqrt{d_h}
}.
\label{eq:head_aggregate_attention}
\end{equation}
The resulting alignment relevance is
\begin{equation}
r_i^{\mathrm{attn}}
=
\max_{j\in\mathcal{I}_{\mathrm{anc}}}
A_{j,i}.
\label{eq:method_attn}
\end{equation}
Thus $r_i^{\mathrm{attn}}$ is a pre-softmax alignment score rather than an attention probability.

\paragraph{Cue fusion.}
Because the cues have different scales, we min--max normalize each visual-token score vector within the current sample.
For $\mathbf{r}\in\mathbb{R}^{N_v}$, define
\begin{equation}
\operatorname{Norm}(\mathbf{r})
=
\frac{\mathbf{r}-\min(\mathbf{r})}
{\max(\mathbf{r})-\min(\mathbf{r})+\epsilon}.
\label{eq:minmax_norm}
\end{equation}
Here $\epsilon=10^{-6}$ is a numerical stabilizer.
Let $\mathbf{r}^{\mathrm{sim}}=[r_1^{\mathrm{sim}},\ldots,r_{N_v}^{\mathrm{sim}}]$ and $\mathbf{r}^{\mathrm{attn}}=[r_1^{\mathrm{attn}},\ldots,r_{N_v}^{\mathrm{attn}}]$.
We average the normalized cues and renormalize the fused scores to restore their sample-wise dynamic range before coverage weighting:
\begin{equation}
\mathbf{u}
=
\operatorname{Norm}
\left(
\frac{
\operatorname{Norm}(\mathbf{r}^{\mathrm{sim}})
+
\operatorname{Norm}(\mathbf{r}^{\mathrm{attn}})
}{2}
\right).
\label{eq:fused_utility}
\end{equation}
Equal weighting introduces no dataset-specific fusion coefficient and is used in all full-QCPruner results. Table~\ref{tab:cue_anchor_ablation} compares it with the single-cue variants.
In what follows, $u_i=[\mathbf{u}]_i$ denotes the query-conditioned utility of visual token $v_i$.

\subsection{Query-Conditioned Population Coverage}
\label{sec:coverage_selection}

QCPruner formulates visual-token selection as query-conditioned population coverage with bilateral contextual weighting.
Each visual token acts as a target, while each selected token serves as a candidate representative. Target utility determines which evidence deserves coverage, representative utility determines which tokens are suitable representatives, and visual affinity links the two roles.

\paragraph{Target--representative utility weighting.}
Before constructing pairwise weights, we apply the fixed backbone-level calibration
\begin{equation}
\hat{u}_i=\rho+(1-\rho)u_i, \qquad \rho\in[0,1].
\label{eq:utility_floor}
\end{equation}
For $0\leq\rho<1$, this transformation preserves the ranking while contracting utility gaps. $\rho=0$ disables the floor, whereas the diagnostic endpoint $\rho=1$ yields visual-only coverage.
We choose $\rho$ once per backbone family on separate calibration data and freeze it for all reported test datasets and budgets. Section~\ref{sec:exp_setup} and Supplementary Section~C.2 give the protocol and sweeps.

We compute nonnegative contextual affinity from visual hidden states:
\begin{equation}
c_{i,j}
=
[\cos(\mathbf{h}^{v}_i,\mathbf{h}^{v}_j)]_+,
\label{eq:visual_affinity}
\end{equation}
where $[x]_+=\max(x,0)$. Negative cosine similarity is treated as absence of coverage rather than as negative evidence.
The query-weighted edge is
\begin{equation}
w_{i,j}=c_{i,j}\hat{u}_i\hat{u}_j.
\label{eq:weighted_coverage_affinity}
\end{equation}
Here $\hat{u}_i$ weights the value of covering target $i$, whereas $\hat{u}_j$ weights the suitability of representative $j$.
Although their product is symmetric for a fixed edge, the two factors play different roles after maximization over selected representatives.

\paragraph{Coverage objective.}
For a selected index set $\mathcal{K}$, token $v_i$ receives coverage
\begin{equation}
\operatorname{cover}_i(\mathcal{K})
=
\max_{j\in\mathcal{K}} w_{i,j},
\qquad
\operatorname{cover}_i(\emptyset)=0.
\label{eq:coverage_state}
\end{equation}
Summing over all candidate tokens gives the bilaterally weighted facility-location objective~\cite{lin2011class,mmtok}:
\begin{equation}
F(\mathcal{K})
=
\sum_{i=1}^{N_v}
\operatorname{cover}_i(\mathcal{K})
=
\sum_{i=1}^{N_v}
\hat u_i\max_{j\in\mathcal{K}} c_{i,j}\hat u_j.
\label{eq:coverage_objective}
\end{equation}
The exact problem is $\mathcal{K}^{*}\in\arg\max_{|\mathcal{K}|=K}F(\mathcal{K})$.
The outer target weight $\hat u_i$ determines how much each target contributes to the objective, while the inner term $c_{i,j}\hat u_j$ determines how well selected token $j$ can represent target $i$.
The maximum prevents multiple representatives from receiving additive credit for the same target, and the outer sum rewards coverage across the full visual population.

\paragraph{Greedy marginal-gain selection.}
Given the current coverage vector, the exact gain of candidate $k$ is
\begin{equation}
\Delta(k\mid\mathcal{K})
=
\sum_{i=1}^{N_v}
\left[
w_{i,k}
-
\operatorname{cover}_i(\mathcal{K})
\right]_+.
\label{eq:coverage_gain}
\end{equation}
For $\mathcal A\subseteq\mathcal B$ and $k\notin\mathcal B$, existing coverage satisfies $\operatorname{cover}_i(\mathcal A)\leq\operatorname{cover}_i(\mathcal B)$, so the target-wise marginal gain of adding $k$ cannot increase as the selected set grows.
Summing these nonincreasing marginals over all targets yields diminishing returns. Together with zero coverage for the empty set and nonnegative edge weights, $F$ is normalized, monotone, and submodular.
The bilateral instance therefore preserves the standard $(1-1/e)$ greedy guarantee for the proxy objective~\cite{nemhauser1978analysis,mmtok}, not downstream accuracy.

Starting from $\mathcal{K}=\emptyset$, we iteratively add the
unselected candidate with the largest marginal gain defined in
Eq.~\eqref{eq:coverage_gain}, update coverage with an elementwise
maximum, and stop once $|\mathcal{K}|=K$, yielding the approximate
solution $\widehat{\mathcal{K}}$.
With a precomputed dense affinity matrix, the direct implementation uses $O(N_v^2)$ memory and $O(KN_v^2)$ worst-case selector time, in addition to affinity construction.
This cost motivates explicit efficiency analysis rather than an unqualified acceleration claim.

\section{Experiments}
\label{sec:experiments}

We test whether bilateral query weighting improves performance retention over visual-only coverage, whether QCPruner outperforms the evaluated complete systems, whether it transfers across architectures and modalities, which components matter, and what costs it incurs.

\begin{figure*}[t]
    \centering
    \includegraphics[width=0.92\textwidth]{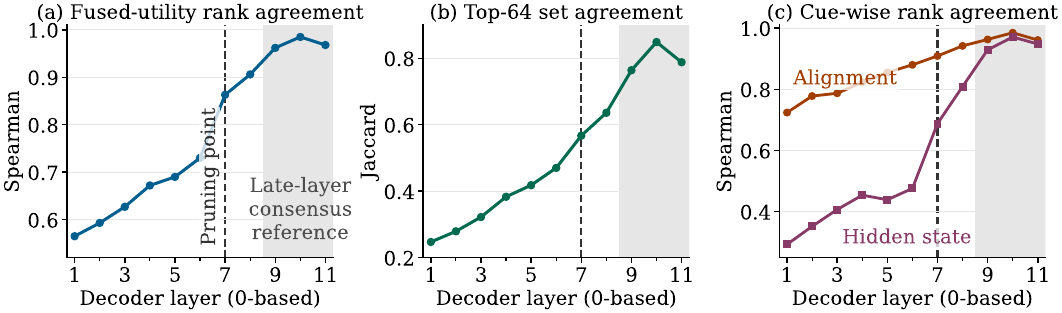}
    \caption{Full layer-wise calibration trajectories on 1,500 MMStar~\cite{mmstar} examples. (a) Fused-utility Spearman agreement, (b) Top-64 Jaccard agreement, and (c) alignment- and hidden-state-cue Spearman agreement with their corresponding per-sample late-layer consensus proxies. The dashed line marks decoder layer 7. The deployed hook removes tokens after layer 7 and before layer 8. The shaded layers 9--11 constitute the same-model late-layer consensus reference and are displayed for completeness rather than treated as independent validation points. All indices are implementation-level and zero-based.}
    \label{fig:app_layer_stability_full}
\end{figure*}

\subsection{Experimental Setup}
\label{sec:exp_setup}

\noindent\textbf{Model architectures.}
We evaluate LLaVA-1.5-7B (576 image tokens), LLaVA-NeXT-7B (up to 2880 high-resolution tokens under native AnyRes preprocessing), LLaVA-Video-7B (64 frames, each with 169 patch tokens), and Qwen2.5-VL-7B (1296 image tokens).
Their retained budgets are 128/64/32, 640/320/160, 4096/2048/1024, and 512/256/128 patch tokens, respectively.
For video, QCPruner solves Eq.~\eqref{eq:coverage_objective} independently within each frame under the same per-frame budget and does not allocate tokens adaptively across frames. All methods preserve frame-grid newline tokens, which are excluded from the reported budgets.
Supplementary Section~A.1 and Supplementary Table~1 summarize the backbone-specific pruning settings.

\noindent\textbf{Evaluation benchmarks.}
For LLaVA image models, we use VQAv2~\cite{vqav2}, GQA~\cite{gqa}, VizWiz~\cite{vizwiz}, ScienceQA-IMG~\cite{scienceqa}, TextVQA~\cite{textvqa}, POPE~\cite{pope}, MME~\cite{mme}, MMBench-EN/CN~\cite{mmbench}, and MMVet~\cite{mmvet}.
The Qwen2.5-VL suite comprises TextVQA, ChartQA~\cite{chartqa}, AI2D~\cite{ai2d}, OCRBench~\cite{ocrbench}, MME, and MMBench-EN/CN. The video suite comprises MVBench~\cite{mvbench}, LongVideoBench~\cite{longvideobench}, and Video-MME~\cite{videomme}.
We follow each benchmark's official split and evaluation metric.
Supplementary Section~A.3 provides additional benchmark details.

\noindent\textbf{Baselines and budget matching.}
We compare with attention-, similarity-, diversity-, coverage-, and conditional-subset-based methods: FastV, SparseVLM, VisionZip, DART, DivPrune, CDPruner, MMTok, and PruneSID~\cite{fastv,sparsevlm,visionzip,dart,divprune,cdpruner,mmtok,prunesid}, plus FastVID~\cite{fastvid} for video.
All baseline and QCPruner scores within each backbone are measured in the same local environment rather than copied from source papers.
Supplementary Sections~A.4 and B report baseline-specific budget details and all compatible results obtained across architectures and budgets.

\noindent\textbf{QCPruner settings.}
Before downstream evaluation, we select the zero-based pruning hook on separate MMStar calibration data using LLaVA-1.5-7B at 64/576 tokens and transfer it across backbones, budgets, and test sets.
Across all 1,500 calibration samples, Figure~\ref{fig:app_layer_stability_full} shows that fused-utility agreement with the same-model layers 9--11 consensus has its largest early increase from layer 6 to 7, with Spearman rising from 0.730 to 0.863 and Top-64 Jaccard from 0.470 to 0.567.
We therefore choose layer 7 as the earliest post-transition stability--efficiency operating point, allowing one additional decoder layer to process the pruned sequence relative to layer 8.
The consensus is an internal proxy rather than independent validation.
Supplementary Section~C.6 provides the metric definitions, complete layer-wise values, and detailed analysis, while Supplementary Section~F provides qualitative examples.

Cue weights are fixed equally, at most six keywords are extracted, and all settings are fixed per model family rather than per benchmark.
We separately calibrate the utility floor $\rho$ outside the reported test suite and then freeze it across downstream datasets and budgets.
The resulting values are $0$ for LLaVA image models, $0.4$ for Qwen2.5-VL, and $0.5$ for LLaVA-Video.
These values are selected from aggregate calibration sweeps rather than claimed as uniquely optimal coefficients. Supplementary Section~C.2 and Table~7 provide the full protocol and results.

Experiments run on an NVIDIA RTX 5880 Ada GPU with 48\,GB memory.
Supplementary Sections~A.5--C provide implementation details, complete results, and calibration analyses.

\subsection{Main Results on LLaVA-1.5}
\label{sec:main_results_llava15}
\begin{table*}[t]
\centering
\small
\setlength{\tabcolsep}{1mm}
\begin{tabular}{lccccccccccc}
\toprule
\textbf{Method} & \textbf{VQAv2} & \textbf{GQA} & \textbf{VizWiz} & \textbf{SQA} & \textbf{TextVQA} & \textbf{POPE} & \textbf{MME} & \textbf{MMB-E} & \textbf{MMB-C} & \textbf{MMVet} & \textbf{Avg. Rel.} \\
\midrule
\rowcolor{tablegray}
\multicolumn{12}{c}{\textit{All 576 Tokens (100\%)}} \\
Full & 78.5 & 61.9 & 50.1 & 69.5 & 58.2 & 85.9 & 1506.5 & 64.7 & 58.1 & 31.3 & 100.0 \\
\midrule
\rowcolor{tablegray}
\multicolumn{12}{c}{\textit{128 Tokens ($\downarrow$77.8\%)}} \\
MMTok & 76.4 & 59.2 & \textbf{53.0} & 68.9 & 56.8 & 86.5 & 1425.8 & 61.0 & 55.5 & 30.8 & 97.9 \\
VisionZip & 75.6 & 57.6 & 52.1 & 68.8 & 56.8 & 83.1 & 1433.3 & 61.3 & 56.7 & \textbf{32.9} & 97.9 \\
DivPrune & 76.0 & 59.4 & 52.8 & 68.5 & 55.9 & 87.0 & 1401.2 & 60.8 & 54.8 & 30.7 & 97.3 \\
CDPruner & 76.6 & 59.6 & 52.7 & 69.0 & 56.1 & \textbf{87.5} & 1426.5 & 62.4 & 55.1 & 30.2 & 97.9 \\
\textbf{QCPruner} & \textbf{77.7} & \textbf{61.2} & 50.9 & \textbf{69.2} & \textbf{57.7} & 86.6 & \textbf{1500.1} & \textbf{63.4} & \textbf{57.6} & 30.5 & \textbf{99.3} \\
\midrule
\rowcolor{tablegray}
\multicolumn{12}{c}{\textit{64 Tokens ($\downarrow$88.9\%)}} \\
MMTok & 75.2 & 58.2 & \textbf{53.8} & 68.8 & 55.8 & 85.6 & 1402.3 & 59.4 & 53.9 & 27.5 & 95.7 \\
VisionZip & 72.4 & 55.1 & 52.9 & 68.9 & 55.4 & 77.0 & 1364.2 & 59.3 & 55.3 & \textbf{31.7} & 94.9 \\
DivPrune & 74.2 & 57.7 & \textbf{53.8} & 67.9 & 54.5 & 85.5 & 1345.0 & 59.1 & 52.3 & 28.6 & 94.8 \\
CDPruner & 75.3 & 58.6 & 53.4 & 68.0 & 55.1 & \textbf{87.5} & 1403.1 & 60.2 & 53.3 & 28.3 & 96.0 \\
\textbf{QCPruner} & \textbf{76.9} & \textbf{60.8} & 50.6 & \textbf{69.5} & \textbf{56.8} & 86.7 & \textbf{1473.2} & \textbf{63.7} & \textbf{56.5} & 29.1 & \textbf{98.2} \\
\midrule
\rowcolor{tablegray}
\multicolumn{12}{c}{\textit{32 Tokens ($\downarrow$94.4\%)}} \\
MMTok & 73.1 & 56.2 & \textbf{54.5} & 68.8 & 53.5 & 85.9 & 1350.1 & 58.1 & 49.3 & 27.0 & 93.4 \\
VisionZip & 67.3 & 51.7 & 52.7 & 68.6 & 53.1 & 68.7 & 1243.8 & 56.8 & 50.2 & 26.3 & 88.5 \\
DivPrune & 71.2 & 54.9 & 53.4 & 68.7 & 52.9 & 81.5 & 1288.0 & 56.8 & 49.1 & 26.8 & 91.4 \\
CDPruner & 73.5 & 56.9 & 53.1 & 69.4 & 53.2 & \textbf{87.7} & 1371.5 & 58.8 & 49.5 & 27.2 & 93.9 \\
\textbf{QCPruner} & \textbf{75.2} & \textbf{59.6} & 50.1 & \textbf{70.0} & \textbf{54.6} & 86.4 & \textbf{1418.6} & \textbf{62.2} & \textbf{55.6} & \textbf{27.4} & \textbf{96.1} \\
\bottomrule
\end{tabular}
\caption{LLaVA-1.5-7B results under common nominal patch-token targets. Avg. Rel. is the equal-weight average of task scores normalized by the corresponding unpruned scores. Bold denotes the best value among the displayed pruning methods. Supplementary Table~3 contains every result obtained for the compatible settings.}
\label{tab:main_llava15}
\end{table*}

At 128, 64, and 32 tokens, QCPruner achieves Avg. Rel. scores of 99.3, 98.2, and 96.1, exceeding the strongest displayed complete-system baseline by 1.4, 2.2, and 2.2 points, respectively.
Against MMTok specifically, QCPruner is higher by 1.4, 2.5, and 2.7 points under the same metric.
QCPruner shows consistent gains on VQAv2, GQA, MME, and MMBench, while QCPruner trails the best displayed result on VizWiz and POPE and at some MMVet budgets.
Thus Table~\ref{tab:main_llava15} supports stronger average downstream performance retention rather than uniform task-wise dominance.

\subsection{Results Across Architectures and Modalities}
\label{sec:generalization}

We further evaluate QCPruner across high-resolution image, video, and Qwen2.5-VL settings.
Table~\ref{tab:generalization_summary} names the strongest evaluated baseline at each budget. Supplementary Tables~4--6 provide the per-benchmark scores, and Avg. Rel. is computed over each backbone's corresponding task set.

\begin{table}[t]
\centering
\small
\setlength{\tabcolsep}{0.35mm}
\begin{tabular}{@{}lllccc@{}}
\toprule
\textbf{Backbone} & \textbf{Target/Full} & \textbf{Baseline} & \textbf{Base.} & \textbf{Ours} & $\boldsymbol{\Delta}$ \\
\midrule
LLaVA-NeXT & 640/2880 & SparseVLM & 97.8 & \textbf{98.1} & +0.3 \\
LLaVA-NeXT & 320/2880 & DART & 95.0 & \textbf{96.6} & +1.6 \\
LLaVA-NeXT & 160/2880 & CDPruner & 92.8 & \textbf{94.9} & +2.1 \\
LLaVA-Video & 4096/10816 & FastVID & 99.4 & \textbf{99.6} & +0.2 \\
LLaVA-Video & 2048/10816 & FastVID & 97.0 & \textbf{98.1} & +1.1 \\
LLaVA-Video & 1024/10816 & FastVID & 94.3 & \textbf{96.3} & +2.0 \\
Qwen2.5-VL & 512/1296 & VisionZip & 98.4 & \textbf{98.5} & +0.1 \\
Qwen2.5-VL & 256/1296 & VisionZip & 92.5 & \textbf{96.7} & +4.2 \\
Qwen2.5-VL & 128/1296 & VisionZip & 82.9 & \textbf{92.8} & +9.9 \\
\bottomrule
\end{tabular}
\caption{All reported budgets beyond LLaVA-1.5. Base. is the Avg. Rel. of the named strongest evaluated baseline at the same nominal patch-token target. $\Delta$ is the absolute percentage-point difference computed before rounding.}
\label{tab:generalization_summary}
\end{table}

The margin over the strongest evaluated baseline generally widens as the token budget tightens (Table~\ref{tab:generalization_summary}). 
Against MMTok, the gap increases most clearly on Qwen2.5-VL, from 2.8 to 11.7 Avg. Rel. points from the loosest to the tightest budget, with consistent gains also observed on LLaVA-1.5 and LLaVA-NeXT.
At the tightest budgets, the gains are also distributed across tasks rather than driven by a single benchmark: QCPruner leads all seven reported Qwen2.5-VL tasks, 
all six video metrics, and seven of ten LLaVA-NeXT tasks while tying one more. Overall, the retention advantage extends across image, video, and model architectures.

\subsection{Retention--Cost Trade-off}
\label{sec:efficiency}

Full QCPruner prioritizes retention: all visual tokens traverse decoder layers 0--7, after which dense affinity is constructed and greedy coverage selection prunes the sequence before layer 8.
QCPruner-Early is a distinct visual-only control before decoder layer 0 and does not use query cues or the full query-conditioned population-coverage objective.

\begin{table}[t]
\centering
\small
\setlength{\tabcolsep}{1mm}
\begin{tabular}{lccccc}
\toprule
\textbf{Method} & \textbf{FLOPs} & \textbf{LLM Pre.} & \textbf{LLM Dec.} & \textbf{Mem.} & \textbf{F1} \\
& \textbf{(T)} & \textbf{(ms)} & \textbf{(ms)} & \textbf{(GB)} & \\
\midrule
\rowcolor{tablegray}
\multicolumn{6}{c}{\textit{All 2880 Tokens (100\%)}} \\
Unpruned & 32.8 & 271.4 & 31.3 & 15.8 & 86.5 \\
\midrule
\rowcolor{tablegray}
\multicolumn{6}{c}{\textit{320 Tokens ($\downarrow$88.9\%)}} \\
FastV & 5.9 & 63.5 & 71.5 & 15.0 & 78.4 \\
SparseVLM & 6.2 & 82.5 & \textbf{22.2} & 20.3 & 82.7 \\
DART & 5.9 & 73.0 & 22.3 & \textbf{14.7} & 83.5 \\
DivPrune & \textbf{5.0} & 41.8 & 22.5 & \textbf{14.7} & 84.1 \\
QCPruner-Early & \textbf{5.0} & \textbf{40.6} & 22.4 & \textbf{14.7} & 84.4 \\
\textbf{QCPruner} & 11.1 & 127.5 & 25.0 & \textbf{14.7} & \textbf{88.4} \\
\bottomrule
\end{tabular}
\caption{Model-trajectory cost and POPE F1 on LLaVA-NeXT-7B at 320/2880 patch tokens, evaluated on the full POPE set (8,900 samples) and measured on an RTX 5880 Ada.}
\label{tab:efficiency}
\end{table}

Relative to unpruned inference, full QCPruner reduces the model trajectory from 32.8 to 11.1 TFLOPs and prefill from 271.4 to 127.5\,ms while raising POPE F1 from 86.5 to 88.4.
It is nevertheless more expensive than DivPrune at the same budget (5.0 TFLOPs and 41.8\,ms), while QCPruner-Early requires 5.0 TFLOPs and 40.6\,ms and reaches 84.4 F1.
QCPruner-Early matches the low model-side cost of early pruning but sacrifices retention, whereas full QCPruner trades additional contextual computation for substantially higher F1.
The LLM prefill timing in Table~\ref{tab:efficiency} includes QCPruner's pruning-hook execution but excludes query-anchor construction.
Supplementary Section~C.1 reports that initialized query-anchor construction averages 1.67\,ms/sample outside the hook.
The current implementation therefore favors performance retention over maximal pruning speed, with repeated dense marginal-gain evaluation as the main optimization target.

\subsection{Ablation Studies}
\label{sec:ablation}

We ablate the key components connecting query conditioning to coverage-based selection. Pruning-layer selection is calibrated separately as described in Section~\ref{sec:exp_setup}.
Cue, anchor, bilateral, and selector ablations use the corresponding seven-task aggregates.

\subsubsection{Query Conditioning and Bilateral Weighting}
\label{sec:cue_ablation}
\label{sec:anchor_ablation}

We first vary the utility estimator under the most aggressive 32-token budget while holding the post-layer-7 hook and population-coverage selector fixed.

\begin{table}[t]
\centering
\small
\setlength{\tabcolsep}{0.2mm}
\begin{tabular}{@{}lcccccccc@{}}
\toprule
\multicolumn{9}{c}{\textit{LLaVA-1.5-7B, Retain 32 Tokens}} \\
\midrule
\textbf{Configuration} & \textbf{SQA} & \textbf{GQA} & \textbf{POPE} & \textbf{MME} & \textbf{Text} & \textbf{M-E} & \textbf{M-C} & \textbf{Rel.} \\
\midrule
Unpruned & 69.5 & 61.9 & 85.9 & 1506.5 & 58.2 & 64.7 & 58.1 & 100.0 \\
\midrule
Sim.+Anc. & 68.4 & 59.2 & \textbf{86.5} & 1413.1 & 53.5 & 62.0 & 54.7 & 95.8 \\
QK+Anc. & 69.0 & 58.7 & 83.4 & 1424.1 & 54.8 & 61.9 & 54.6 & 95.6 \\
Fusion+Prompt & 68.2 & 58.9 & 86.1 & \textbf{1425.5} & \textbf{55.1} & 62.0 & 55.0 & 96.2 \\
Fusion+Anc. & \textbf{70.0} & \textbf{59.6} & 86.4 & 1418.6 & 54.6 & \textbf{62.2} & \textbf{55.6} & \textbf{96.8} \\
\bottomrule
\end{tabular}
\caption{Cue and query-anchor ablations on seven tasks. Anc. denotes matched anchors. Text/M-E/M-C abbreviate TextVQA/MMBench-EN/MMBench-CN. Rel. is seven-task Avg. Rel. Bold marks the best pruned result.}
\label{tab:cue_anchor_ablation}
\end{table}

At 32 tokens, Fusion+Anc., QK+Anc., and Sim.+Anc. reach 96.8, 95.6, and 95.8 Avg. Rel., respectively (Table~\ref{tab:cue_anchor_ablation}).
At 64 tokens, the corresponding values are 98.6, 98.1, and 97.4 (Supplementary Section~C.7).
Fusion and matched anchors thus give aggregate gains at both budgets, while individual cues remain strongest on some tasks. Supplementary Section~C.6 reports cue-specific layer stability.

Table~\ref{tab:bilateral_utility_main} shows that both one-sided factors improve Avg. Rel. over visual-only coverage in all three settings, with representative weighting providing the stronger one-sided signal.
The bilateral form raises LLaVA from 98.0 to 98.6 and raises Qwen from 92.6 to 92.8 at 128 tokens, while the two Qwen variants both reach 96.7 at 256 tokens at the reported precision.
Its incremental gain over representative-only weighting is therefore modest and setting-dependent.
Task-level results further sharpen this distinction: bilateral weighting outperforms representative-only weighting on five of seven LLaVA tasks and five of seven Qwen-128 tasks. At Qwen-256, the two variants each lead on three tasks and tie on one.
Thus, bilateral weighting yields setting-dependent task-level trade-offs rather than a universal per-task gain (Supplementary Section~C.3).

\begin{table}[t]
\centering
\small
\setlength{\tabcolsep}{1.2mm}
\begin{tabular}{lccc}
\toprule
& \multicolumn{2}{c}{\textbf{Qwen2.5-VL}} & \textbf{LLaVA-1.5} \\
\cmidrule(lr){2-3}\cmidrule(l){4-4}
\textbf{Edge weight} & \textbf{128/1296} & \textbf{256/1296} & \textbf{64/576} \\
\midrule
$c_{i,j}$ & 87.2 & 93.4 & 95.7 \\
$c_{i,j}\hat u_i$ & 87.9 & 94.3 & 97.3 \\
$c_{i,j}\hat u_j$ & 92.6 & \textbf{96.7} & 98.0 \\
$c_{i,j}\hat u_i\hat u_j$ & \textbf{92.8} & \textbf{96.7} & \textbf{98.6} \\
\bottomrule
\end{tabular}
\caption{Bilateral utility ablations across three settings (seven-task Avg. Rel.). Supplementary Section~C.3 reports per-task scores.}
\label{tab:bilateral_utility_main}
\end{table}

Across two to eight keywords, Avg. Rel. ranges from 97.8 to 98.6 and remains at 98.6 at the reported precision for four, six, and eight keywords.
Replacing SparseVLM's original text conditioning with the same keyword-matched anchors improves all seven tasks and raises Avg. Rel. from 90.9 to 91.9 (Supplementary Sections~C.4 and~D).
\subsubsection{Effect of Selection Strategy}
\label{sec:selector_ablation}

We next isolate the subset objective by fixing candidates, anchors, fused query utility, the post-layer-7 hook, utility floor, and budget within each setting.
The alternatives are Top-$K$, visual-token maximal marginal relevance (MMR)~\cite{carbonell1998mmr} with $\lambda_{\mathrm{MMR}}=0.5$, visual-only facility location, a conditional-DPP/Fast-MAP backend, an MMTok-style additive-coverage backend~\cite{mmtok}, and QCPruner.
The DPP and MMTok-style entries are controlled backend transplants rather than complete-system results.
The MMTok-style backend follows the official MMTok objective form and method-specific hyperparameters without retuning.

\begin{table}[t]
\centering
\small
\setlength{\tabcolsep}{1.0mm}
\begin{tabular}{lccc}
\toprule
& \multicolumn{2}{c}{\textbf{Qwen2.5-VL}} & \textbf{LLaVA-1.5} \\
\cmidrule(lr){2-3}\cmidrule(l){4-4}
\textbf{Selector} & \textbf{128/1296} & \textbf{256/1296} & \textbf{64/576} \\
\midrule
Top-$K$ & 69.6 & 78.6 & 96.5 \\
MMR & 88.7 & 93.5 & 98.3 \\
Visual-only Cov. & 87.2 & 93.4 & 95.7 \\
CDPruner-DPP & 89.0 & 93.7 & \textbf{98.6} \\
MMTok-style & 89.7 & 95.1 & 97.8 \\
QCPruner & \textbf{92.8} & \textbf{96.7} & \textbf{98.6} \\
\bottomrule
\end{tabular}
\caption{Controlled seven-task selector ablations (Avg. Rel.). Upstream signals and budget are fixed within each column. CDPruner-DPP and MMTok-style denote backend transplants rather than complete systems.}
\label{tab:qwen_selector_ablation}
\end{table}

Table~\ref{tab:qwen_selector_ablation} shows that QCPruner exceeds the CDPruner-DPP backend by 3.8 and 3.0 points and the MMTok-style backend by 3.1 and 1.6 points at Qwen-128 and Qwen-256, respectively.
On LLaVA, QCPruner and CDPruner-DPP both reach 98.6 at the reported precision.
QCPruner exceeds the MMTok-style backend by 0.8 points in this setting.
Relative to visual-only coverage, it gains 5.6, 3.3, and 2.9 points at Qwen-128, Qwen-256, and LLaVA-64, respectively.
At Qwen-128, QCPruner leads all seven reported tasks. At Qwen-256, it leads five, while MMTok-style and Top-$K$ lead OCRBench and MMBench-EN, respectively.
On LLaVA, the task-level leaders are distributed across QCPruner, CDPruner-DPP, and MMR on three, two, and two tasks, respectively (Supplementary Section~C.5).
Query weighting therefore improves performance retention over visual-only coverage in all three settings, while the relative ordering of selector objectives varies across settings.
Because the columns also differ in task set, token ratio, and $\rho$, these results do not establish universal selector dominance or superiority over the complete CDPruner or MMTok pipelines.
Supplementary Section~C.5 reports task-level results.

\section{Conclusion}
QCPruner introduces query-conditioned bilateral weighting into coverage-based visual-token pruning, allowing contextual query evidence to determine both what deserves coverage and what can serve as a representative.
Across four MLLM backbones and twelve token budgets, it achieves the highest observed Avg. Rel. among the complete systems evaluated in our pipeline, including matched comparisons with MMTok on three image backbones.
Controlled ablations further show that query weighting consistently improves performance retention over visual-only coverage, with representative utility providing the stronger one-sided signal and the additional benefit of bilateral weighting varying across settings.
These results support query-conditioned population coverage as an effective approach to performance retention under tight visual-token budgets.
The current design still incurs the cost of several dense decoder layers before pruning and dense greedy selection, leaving efficiency, layer transfer, grounding, multilingual robustness, and broader failure analysis as directions for future work.

\bibliography{references}
\clearpage
\appendix
This supplementary material is organized as follows.
Appendix~\ref{app:exp_details} provides experimental details.
Appendix~\ref{app:complete_benchmark_tables} reports complete benchmark-level results.
Appendix~\ref{app:additional_ablation} provides additional ablations, including the pruning-layer stability analysis under separate calibration.
Appendix~\ref{app:transferability} studies the transferability of keyword-matched query anchors.
Appendix~\ref{app:limitations} discusses limitations and broader impact.
Appendix~\ref{app:qualitative_examples} provides the qualitative motivation examples omitted from the main paper due to space constraints.
Throughout, full QCPruner denotes query-conditioned population coverage, whereas QCPruner-Early is a separate visual-only control.

\section{Additional Experimental Details}
\label{app:exp_details}

\subsection{Model Architectures and Pruning Settings}
\label{app:model_details}

Table~\ref{tab:app_model_settings} summarizes the evaluated backbones, patch-token targets, and full-QCPruner pruning layer and backbone-level hyperparameters.
Layer indices throughout the paper follow the implementation-level zero-based decoder indexing: layer $\ell$ supplies the contextual states used for pruning, which is performed after layer $\ell$ and before layer $\ell+1$.
For full QCPruner, the post-layer-7 hook is shared across backbones, while equal-weight cue fusion and the selected utility-floor coefficient are fixed across datasets and budgets within each backbone. QCPruner-Early is specified separately in Appendix~\ref{app:early_coverage}.

For Qwen2.5-VL and LLaVA-Video, we apply a utility-floor calibration to mitigate overly concentrated utility distributions during population-coverage selection.
Given the fused utility score $u_i$, the calibrated utility is defined as
\begin{equation}
\hat{u}_i = \rho + (1-\rho)u_i
\label{eq:app_utility_floor}
\end{equation}
For $0<\rho<1$, this transformation preserves the ranking induced by $u_i$ while contracting utility differences, preventing moderately relevant tokens from being excessively downweighted during coverage selection.
When $\rho=0$, the calibration is disabled.
$\rho=1$ makes every calibrated utility equal to one and is used only as a visual-only diagnostic endpoint.
We set $\rho=0$ for LLaVA image backbones, $\rho=0.4$ for Qwen2.5-VL, and $\rho=0.5$ for LLaVA-Video, selected once on separate calibration data and then frozen across downstream datasets and budgets (Appendix~\ref{app:rho_ablation}).

For LLaVA-Video-7B, QCPruner performs selection independently within each frame under a fixed per-frame patch-token budget. All newline tokens separating frame-level visual grids are preserved and excluded from the reported pruning budget.

\subsection{QCPruner-Early Variant}
\label{app:early_coverage}

QCPruner-Early is a separate visual-only pre-decoder control that shortens the token sequence processed by the LLM.
It operates on visual-token positions in the assembled input-embedding tensor immediately before decoder layer 0 and does not use the full query-conditioned objective.
Let $\mathbf{E}_{\mathrm{in}}$ denote the assembled input-embedding tensor, and let
$\mathbf{x}_i=\operatorname{float32}(\mathbf{E}_{\mathrm{in}}[p_i,:])\in\mathbb{R}^{d}$
for candidate position $p_i$.

\paragraph{Visual affinity and norm prior.}
The variant first computes a nonnegative cosine-affinity matrix:
\begin{equation}
c^{\mathrm{early}}_{i,j}
=
[\cos(\mathbf{x}_i,\mathbf{x}_j)]_+,
\qquad
c^{\mathrm{early}}_{i,i}=1.
\label{eq:app_early_affinity}
\end{equation}
The diagonal is explicitly reset to one after clipping.
Because query-conditioned contextual cues are not yet available from decoder-layer interactions at this early insertion point, we use the L2 norm of each raw visual embedding as a lightweight candidate prior:
\begin{equation}
n_i = \|\mathbf{x}_i\|_2
\label{eq:app_early_norm}
\end{equation}
For each sample, the implementation uses the 5th and 95th percentiles $Q_{.05}$ and $Q_{.95}$ of the embedding norms to construct a robust min--max score.
We clip the robust min--max score as
\begin{equation}
\tilde{n}_i
=
\operatorname{clip}
\left(
\frac{n_i - Q_{.05}}
{Q_{.95} - Q_{.05} + \epsilon},
0,1
\right),
\label{eq:app_early_norm_score}
\end{equation}
where $\epsilon=10^{-6}$, and define
\begin{equation}
a_i = 1 + \beta \tilde{n}_i,
\label{eq:app_early_weight}
\end{equation}
with $\beta=0.20$ in all reported QCPruner-Early experiments, giving $a_i\in[1,1.2]$.
Neither $a_i$ nor $c^{\mathrm{early}}_{i,j}$ uses text- or query-derived information.

\paragraph{Coverage-inspired selection.}
For $N_v>K$, the first selected token combines aggregate visual affinity with the norm prior:
\begin{equation}
j_1
=
\arg\max_j
\left(
a_j
\sum_{i=1}^{N_v}
c^{\mathrm{early}}_{i,j}
\right).
\label{eq:app_early_first}
\end{equation}
For a selected set $\mathcal{K}$, define the current coverage of token $v_i$ as
\begin{equation}
\operatorname{cover}^{\mathrm{early}}_i(\mathcal{K})
=
\max_{j\in\mathcal{K}}
c^{\mathrm{early}}_{i,j}.
\label{eq:app_early_coverage}
\end{equation}
Each subsequent step scores a remaining candidate by its uncoveredness relative to the selected set, weighted by the norm prior:
\begin{equation}
j_t
=
\arg\max_{j\notin\mathcal{K}}
\left[
1-
\operatorname{cover}^{\mathrm{early}}_j(\mathcal{K})
\right]_+
a_j .
\label{eq:app_early_select}
\end{equation}
After adding $j_t$, the vector is updated by
\begin{equation}
\operatorname{cover}^{\mathrm{early}}_i(\mathcal{K}\cup\{j_t\})
=
\max
\left(
\operatorname{cover}^{\mathrm{early}}_i(\mathcal{K}),
c^{\mathrm{early}}_{i,j_t}
\right).
\label{eq:app_early_update}
\end{equation}
The procedure selects exactly $K$ visual tokens when $N_v>K$. Otherwise, the visual sequence is left unchanged.

Importantly, the score above uses candidate $j$'s own uncoveredness rather than the summed facility-location marginal gain over all targets used by full QCPruner.
QCPruner-Early is therefore a coverage-inspired, norm-weighted self-novelty heuristic and does not inherit the full objective's approximation guarantee.
Constructing the dense affinity matrix requires $O(N_v^2d)$ time and $O(N_v^2)$ memory.
Given this matrix, column summation and $K$ vector updates require $O(N_v^2+KN_v)$ time.
Its low model-side cost in the main-paper retention--cost comparison comes from pruning before all LLM blocks rather than eliminating dense-affinity preprocessing.

\begin{table*}[t]
\centering
\small
\setlength{\tabcolsep}{1.2mm}
\begin{tabular}{@{}lccc@{}}
\toprule
\textbf{Backbone} & \textbf{Patch tokens} & \textbf{Prune after layer} & $\boldsymbol{\rho}$ \\
& \textbf{full $\rightarrow$ retained} & & \\
\midrule
LLaVA-1.5-7B & $576 \rightarrow 128/64/32$ & 7 & 0.0 \\
LLaVA-NeXT-7B & $2880^{\dagger} \rightarrow 640/320/160$ & 7 & 0.0 \\
Qwen2.5-VL-7B & $1296 \rightarrow 512/256/128$ & 7 & 0.4 \\
LLaVA-Video-7B & $10816 \rightarrow 4096/2048/1024$ & 7 & 0.5 \\
\bottomrule
\end{tabular}
\caption{Backbone-specific pruning settings and nominal patch-token targets. 
$^{\dagger}$ Maximum full patch-token count, while the actual count is input-dependent.
The full LLaVA-Video input contains $64\times169$ patch tokens. 
QCPruner-Early operates before decoder layer 0 and does not use $\rho$. }
\label{tab:app_model_settings}
\end{table*}

LLaVA-1.5, Qwen2.5-VL, and LLaVA-Video use fixed visual-token layouts within each backbone. In contrast, LLaVA-NeXT retains its native AnyRes multi-scale preprocessing, resulting in input-dependent full-token counts. This setup evaluates transfer across controlled backbone settings while separately examining robustness to native multi-scale inputs.

\subsection{Evaluation Benchmarks}
\label{app:dataset_details}

We provide detailed descriptions of the evaluation benchmarks used in our experiments in the following subsections.
All evaluations follow the official protocol of each benchmark, and we report the corresponding task-specific metric.
For a method $m$, we compute average relative performance retention as
\[
\operatorname{Avg.Rel.}(m)=\frac{100}{T}\sum_{t=1}^{T}
\frac{s_{m,t}}{s_{\mathrm{full},t}},
\]
where $s_{m,t}$ and $s_{\mathrm{full},t}$ are the method and unpruned scores on benchmark $t$, respectively, and $T$ is the number of top-level benchmarks in the corresponding table.
Thus, Avg. Rel. gives every benchmark equal weight despite differences in its native score scale.
For the video results, the three top-level terms are MVBench test, LongVideoBench validation, and Video-MME without subtitles. The Short/Medium/Long Video-MME columns are diagnostic breakdowns and are not averaged again.

\subsubsection{General Image Benchmarks}

\paragraph{VQAv2.}
VQAv2~\cite{vqav2} evaluates general visual question answering over natural images, covering objects, attributes, scenes, and simple relations.
For visual token pruning, it provides a broad test of whether the retained visual subset preserves task-relevant evidence across diverse question types.
We use the test-dev split and report the official VQA accuracy, which measures agreement with the human reference answers.

\paragraph{GQA.}
GQA~\cite{gqa} focuses on compositional visual reasoning over objects, attributes, and relationships.
Its emphasis on structured and relational reasoning makes it useful for evaluating whether pruning preserves both salient visual evidence and the supporting context required for compositional answers.
We use the balanced test-dev split and report answer accuracy, the benchmark's direct measure of compositional question-answering performance.

\paragraph{VizWiz.}
VizWiz~\cite{vizwiz} contains real-world images and questions collected from blind users.
The images are often blurry, poorly framed, or visually ambiguous, while the questions can be short or highly referential.
These characteristics make VizWiz challenging for query-aware pruning because some queries, such as ``What is this?'', provide limited semantic guidance for identifying relevant visual evidence.
We use the test split and report the official VQA accuracy against crowdsourced reference answers.

\paragraph{ScienceQA-IMG.}
ScienceQA-IMG~\cite{scienceqa} evaluates multimodal science reasoning with image context.
Its questions often require combining visual evidence with commonsense or domain knowledge.
It therefore evaluates whether pruning preserves the visual evidence needed for multimodal reasoning.
We use the image-context subset of the test split and report multiple-choice accuracy, which directly measures the fraction of science questions answered correctly.

\paragraph{POPE.}
POPE~\cite{pope} evaluates object hallucination in multimodal models by asking whether specific objects are present in an image.
It is relevant to query-aware token pruning because correct predictions depend on preserving object-level evidence for the queried category.
POPE therefore evaluates performance retention on hallucination-sensitive object-presence decisions rather than serving as a direct token-grounding metric.
We use the test split and report F1, which balances precision and recall for the binary object-presence decisions.

\paragraph{MME.}
MME~\cite{mme} is a comprehensive benchmark covering both perception and cognition.
Its perception tasks include object existence, counting, position, color, OCR, and fine-grained recognition, while its cognition tasks involve commonsense reasoning, calculation, translation, and code-related reasoning.
Its broad task coverage provides a comprehensive evaluation of performance retention under visual token pruning.
We report the official aggregate MME score over the perception and cognition evaluations, whose task-wise scoring rewards both correct positive and negative decisions.
For LLaVA-1.5 and LLaVA-NeXT, we report the official MME perception score.
For Qwen2.5-VL, we report the sum of the perception and cognition scores, following the corresponding backbone-specific evaluation pipeline.
The task-wise scoring rewards both correct positive and negative decisions.
Because the evaluation scope differs, raw MME scores should be compared only within the same backbone and evaluation scope.

\paragraph{MMBench-EN and MMBench-CN.}
MMBench~\cite{mmbench} evaluates broad multimodal capabilities using multiple-choice questions.
We report both the English and Chinese splits to broaden language coverage, without treating them as a controlled cross-lingual robustness test based on translated or paraphrased versions of the same prompts.
For both official evaluation splits, we report CircularEval accuracy, which reduces option-position bias by requiring consistency across circularly permuted choices.

\paragraph{MMVet.}
MMVet~\cite{mmvet} evaluates integrated multimodal capabilities, including recognition, OCR, knowledge, language generation, spatial awareness, and mathematics.
Its questions often require combining multiple skills rather than relying on a single visual cue.
This makes MMVet challenging for aggressive pruning, as removing complementary visual evidence can degrade performance on multi-skill reasoning tasks.
We report the official LLM-judged MM-Vet score for open-ended responses, which evaluates correctness across the benchmark's integrated capabilities.

\subsubsection{Text-Oriented Image Benchmarks}

\paragraph{TextVQA.}
TextVQA~\cite{textvqa} evaluates a model's ability to read and reason over scene text in natural images.
Because answers often depend on fine-grained OCR evidence occupying small image regions, the benchmark is sensitive to whether pruning preserves localized text information together with its supporting visual context.
We use the validation split and report the official VQA accuracy against human reference answers.

\paragraph{ChartQA.}
ChartQA~\cite{chartqa} focuses on question answering over charts.
It requires visual parsing of chart elements together with logical or arithmetic reasoning over values, labels, and trends.
For visual token pruning, ChartQA is challenging because small textual or graphical elements may contain evidence critical to answering the question.
We use the test split and report relaxed accuracy, which accommodates the benchmark's prescribed numerical tolerance while still requiring correct non-numerical answers.

\paragraph{AI2D.}
AI2D~\cite{ai2d} evaluates diagram understanding using science-related diagrams and multiple-choice questions.
Compared with natural images, diagrams contain structured visual elements, symbolic labels, and spatial relations.
For visual token pruning, AI2D tests whether sparse but task-relevant diagram elements are preserved.
We use the test split and report multiple-choice accuracy, the direct fraction of diagram questions answered correctly.

\paragraph{OCRBench.}
OCRBench~\cite{ocrbench} evaluates OCR-oriented multimodal understanding across diverse text-centric visual tasks.
It covers text recognition, scene-text VQA, document-oriented VQA, key information extraction, and handwritten expression recognition.
Because OCR evidence is often fine-grained and spatially localized, OCRBench is particularly sensitive to aggressive visual token pruning.
We use the official 1,000-question evaluation set and report the aggregate number of correctly answered items across its five task groups. The maximum is 1,000 and higher is better.

\subsubsection{Video Benchmarks}

\paragraph{MVBench.}
MVBench~\cite{mvbench} evaluates temporal understanding in multimodal video models.
It includes tasks involving actions, events, temporal order, and dynamic scene changes.
For visual token pruning, MVBench tests whether frame-level evidence needed for temporal reasoning is preserved.
We use the test split and report mean multiple-choice accuracy across its 20 tasks.

\paragraph{LongVideoBench.}
LongVideoBench~\cite{longvideobench} evaluates long-form video understanding.
It requires models to retrieve and reason over information distributed across extended temporal contexts.
For visual token pruning, LongVideoBench is challenging because aggressive compression may remove sparse but task-relevant temporal evidence.
We use the validation split and report multiple-choice accuracy, which directly measures long-video question-answering correctness.

\paragraph{Video-MME.}
Video-MME~\cite{videomme} is a comprehensive video understanding benchmark covering videos of varying durations and reasoning categories.
We use the no-subtitle setting to focus evaluation on visual evidence from video frames.
For visual token pruning, Video-MME evaluates whether query-relevant temporal and visual evidence is preserved under large-scale token reduction.
We report multiple-choice accuracy in the no-subtitle setting, both overall and for the official short-, medium-, and long-duration subsets.
The overall score is used as the top-level Video-MME term in Avg. Rel.
The duration-specific columns are diagnostic breakdowns and are not averaged again.

\subsection{Comparison Methods}
\label{app:comparison_methods}

We report compatible training-free visual token pruning and compression baselines spanning attention-based pruning, text-guided pruning, visual redundancy reduction, and diversity-oriented token selection.

\paragraph{FastV.}
FastV~\cite{fastv} is a training-free pruning method that removes visual tokens according to attention-derived importance.
It serves as a representative baseline for visual token pruning within the language model.

\paragraph{SparseVLM.}
SparseVLM~\cite{sparsevlm} uses text-related attention signals to guide visual token sparsification.
It serves as a representative baseline for text-guided visual token pruning.

\paragraph{VisionZip.}
VisionZip~\cite{visionzip} compresses visual tokens by combining dominant and contextual tokens to preserve both salient and contextual visual information.
It serves as a representative baseline for visual-side token compression.

\paragraph{DART.}
DART~\cite{dart} focuses on reducing visual-token redundancy, treating token duplication as a key source of inefficiency in visual token pruning.
It serves as a representative redundancy-reduction baseline.

\paragraph{DivPrune.}
DivPrune~\cite{divprune} performs diversity-oriented visual token pruning to preserve diverse visual information in the retained subset.
It provides a direct comparison with diversity-based subset selection that does not explicitly condition on the current query.

\paragraph{CDPruner.}
CDPruner~\cite{cdpruner} performs visual token pruning through conditional-diversity modeling with a DPP-style subset selection objective.
The CDPruner rows in the complete benchmark tables refer to the full method and are distinct from the controlled CDPruner-DPP selector transplant in Appendix~\ref{app:dpp_transplant}, 
where CDPruner's native relevance input is replaced with QCPruner-derived query utility.

\paragraph{MMTok.}
MMTok~\cite{mmtok} is closely related to QCPruner in its use of facility-location-style subset selection.
Its pre-decoder objective combines text--vision and vision--vision facility-location terms, yielding a monotone submodular objective optimized by standard greedy selection.
In contrast, QCPruner derives query-conditioned visual-token utility from keyword-matched contextual MLLM states at the pruning layer and uses it to modulate both target importance and representative suitability within visual-population coverage.
We locally evaluate MMTok at the same nominal patch-token budgets on LLaVA-1.5, LLaVA-NeXT, and Qwen2.5-VL.
The controlled selector study in Appendix~\ref{app:dpp_transplant} additionally includes an \emph{MMTok-style} objective-level backend transplant on Qwen2.5-VL and LLaVA-1.5. This entry is distinct from the complete MMTok system reported in the benchmark tables.

\paragraph{PruneSID.}
PruneSID~\cite{prunesid} is a recent training-free visual token pruning method included for additional comparison under matched nominal token budgets.

\paragraph{FastVID.}
For video experiments, we additionally compare with FastVID~\cite{fastvid}, a video-oriented pruning baseline evaluated under matched total token budgets.

\subsection{Implementation and Evaluation Protocol}
\label{app:evaluation_protocol}
\label{app:implementation}

QCPruner uses equal-weight cue fusion together with the backbone-specific pruning layer and utility-floor coefficient $\rho$ listed in Table~\ref{tab:app_model_settings}, whereas QCPruner-Early follows Appendix~\ref{app:early_coverage}.

Keyword-matched query anchors are constructed using KeyBERT v0.9.0~\cite{keybert} with the model2vec v0.8.1 StaticModel \texttt{minishlab/potion-base-8M}.\footnote{\url{https://huggingface.co/minishlab/potion-base-8M}}
Candidate extraction uses unigram range $(1,1)$, KeyBERT's default English stop-word filtering, and at most six ranked keywords, with optional MMR/diversity reranking disabled.
Each returned keyword is matched back to all corresponding token spans in the original prompt, and the matched tokens are used as textual anchors for visual utility estimation.
If no keyword match is found, we fall back to valid non-template text tokens after excluding special tokens and visual placeholders.

For both LLaVA-1.5 and high-resolution LLaVA-NeXT image experiments, we use the official LLaVA codebase.\footnote{\url{https://github.com/haotian-liu/LLaVA}}
For LLaVA-Video experiments, we use the LLaVA-NeXT codebase\footnote{\url{https://github.com/LLaVA-VL/LLaVA-NeXT}} for model implementation and multimodal preprocessing and lmms-eval\footnote{\url{https://github.com/EvolvingLMMs-Lab/lmms-eval}} for benchmark evaluation.
For Qwen2.5-VL, we use the official model implementation and lmms-eval for evaluation.
All benchmarks use deterministic greedy decoding with sampling
disabled and official evaluation settings. Each benchmark score
is obtained from one evaluation pass, while latency is averaged
over a single pass through all 8,900 POPE examples. We do not
report repeated-run uncertainty or statistical significance tests.

Experiments are conducted on a single NVIDIA RTX 5880 Ada GPU with 48\,GB of memory and an Intel Core i9-14900K CPU under Ubuntu 22.04 and Linux 6.8.0.
Because different backbone families rely on different official codebases and software dependencies, we use separate environments for each family.
Table~\ref{tab:app_env} summarizes the corresponding software environments.

\begin{table}[t]
\centering
\small
\setlength{\tabcolsep}{0.8mm}
\begin{tabular}{@{}lp{0.70\columnwidth}@{}}
\toprule
\textbf{Family} & \textbf{Family-specific software stack} \\
\midrule
LLaVA Image & Python 3.10.20; PyTorch 2.1.2; CUDA 11.8; cuDNN 8.7.0; Transformers 4.37.2; Accelerate 0.21.0; official LLaVA evaluation. \\
LLaVA-Video & Python 3.10.20; PyTorch 2.1.2+cu121; CUDA 12.1; cuDNN 8.9.0; Transformers 4.40.0.dev0; Accelerate 1.13.0; lmms-eval 0.7.1. \\
Qwen2.5-VL & Python 3.12.13; PyTorch 2.11.0+cu128; CUDA 12.8; cuDNN 9.19.0; Transformers 5.5.4; Accelerate 1.13.0; lmms-eval 0.7.1. \\
\bottomrule
\end{tabular}
\caption{Family-specific software environments. Hardware and operating-system details shared by all families are stated in the text.}
\label{tab:app_env}
\end{table}

\section{Complete Benchmark Tables}
\label{app:complete_benchmark_tables}

The main paper reports compact tables for readability.
Here we provide the complete benchmark-wise results obtained for all compatible baseline, architecture, and retained-token combinations.
Tables~\ref{tab:app_llava15_full}--\ref{tab:app_qwen_full} report the full results for LLaVA-1.5-7B, LLaVA-NeXT-7B, LLaVA-Video-7B, and Qwen2.5-VL-7B under their respective retained-token budgets.
Overall, QCPruner achieves strong average performance retention across the evaluated complete systems, with larger gains generally emerging under tighter token budgets.

\paragraph{Result overview.}
On LLaVA-1.5, QCPruner achieves the highest Avg. Rel. at all three budgets, including 96.1\% at 32 tokens, 2.2 points above the strongest evaluated baseline, although some individual tasks still show regressions.
On LLaVA-NeXT, the gain over the strongest evaluated baseline increases from +0.3 points at 640 tokens to +2.1 points at 160 tokens.
For video, QCPruner remains competitive at 4096 tokens and achieves the highest average retention at 2048 and 1024 tokens among the evaluated methods under matched token budgets.
Adaptive cross-frame allocation is not evaluated.
On Qwen2.5-VL, QCPruner and VisionZip are nearly tied at 512 tokens, while QCPruner exceeds the strongest evaluated baseline by 4.2 and 9.9 points at 256 and 128 tokens, respectively.
Across the three image backbones, QCPruner also achieves higher Avg. Rel. than MMTok at every matched budget, with the performance gap widening under tighter LLaVA-NeXT and Qwen2.5-VL budgets.

\begin{table*}[t]
\centering
\small
\setlength{\tabcolsep}{1mm}
\begin{tabular}{lccccccccccc}
\toprule
\textbf{Method} 
& \textbf{VQAv2} 
& \textbf{GQA} 
& \textbf{VizWiz} 
& \textbf{SQA} 
& \textbf{TextVQA} 
& \textbf{POPE} 
& \textbf{MME} 
& \textbf{MMB-E} 
& \textbf{MMB-C} 
& \textbf{MMVet} 
& \textbf{Avg. Rel.} \\
\midrule

\rowcolor{tablegray}
\multicolumn{12}{c}{\textit{All 576 Tokens (100\%)}} \\
LLaVA-1.5 
& 78.5 & 61.9 & 50.1 & 69.5 & 58.2 & 85.9 & 1506.5 & 64.7 & 58.1 & 31.3 & 100.0 \\

\rowcolor{tablegray}
\multicolumn{12}{c}{\textit{128 Tokens ($\downarrow$77.8\%)}} \\
FastV 
& 73.2 & 55.4 & 51.4 & 68.1 & 56.4 & 72.3 & 1442.1 & 61.2 & 56.3 & 30.0 & 94.7 \\

SparseVLM 
& 75.3 & 59.4 & 50.1 & 68.6 & 56.7 & 79.6 & 1292.7 & \textbf{63.8} & \textbf{57.9} & 29.1 & 95.8 \\

VisionZip 
& 75.6 & 57.6 & 52.1 & 68.8 & 56.8 & 83.1 & 1433.3 & 61.3 & 56.7 & \textbf{32.9} & 97.9 \\

DART 
& 76.0 & 58.8 & 51.6 & \textbf{69.2} & 56.5 & 80.2 & 1485.5 & 62.5 & 57.4 & 29.2 & 97.2 \\

DivPrune 
& 76.0 & 59.4 & 52.8 & 68.5 & 55.9 & 87.0 & 1401.2 & 60.8 & 54.8 & 30.7 & 97.3 \\

CDPruner 
& 76.6 & 59.6 & 52.7 & 69.0 & 56.1 & \textbf{87.5} & 1426.5 & 62.4 & 55.1 & 30.2 & 97.9 \\

PruneSID 
& 75.4 & 58.1 & 52.0 & 68.0 & 54.4 & 84.7 & 1416.0 & 61.4 & 56.2 & 30.1 & 96.5 \\

MMTok
& 76.4 & 59.2 & 53.0 & 68.9 & 56.8 & 86.5 & 1425.8 & 61.0 & 55.5 & 30.8 & 97.9 \\

QCPruner-Early 
& 76.1 & 59.2 & \textbf{53.2} & 68.6 & 56.6 & 86.9 & 1383.9 & 60.6 & 54.7 & 30.6 & 97.3 \\

\textbf{QCPruner} 
& \textbf{77.7} & \textbf{61.2} & 50.9 & \textbf{69.2} & \textbf{57.7} & 86.6 & \textbf{1500.1} & 63.4 & 57.6 & 30.5 & \textbf{99.3} \\

\rowcolor{tablegray}
\multicolumn{12}{c}{\textit{64 Tokens ($\downarrow$88.9\%)}} \\
FastV 
& 66.3 & 51.6 & 51.5 & 67.3 & 54.7 & 59.5 & 1246.0 & 57.3 & 50.3 & 26.7 & 87.4 \\

SparseVLM 
& 70.2 & 53.7 & 50.1 & \textbf{69.6} & 53.4 & 77.4 & 1290.1 & 59.3 & 52.4 & 24.9 & 90.5 \\

VisionZip 
& 72.4 & 55.1 & 52.9 & 68.9 & 55.4 & 77.0 & 1364.2 & 59.3 & 55.3 & \textbf{31.7} & 94.9 \\

DART 
& 72.7 & 56.2 & 51.5 & 68.7 & 54.3 & 74.1 & 1408.6 & 60.9 & 53.8 & 26.7 & 93.0 \\

DivPrune 
& 74.2 & 57.7 & \textbf{53.8} & 67.9 & 54.5 & 85.5 & 1345.0 & 59.1 & 52.3 & 28.6 & 94.8 \\

CDPruner 
& 75.3 & 58.6 & 53.4 & 68.0 & 55.1 & \textbf{87.5} & 1403.1 & 60.2 & 53.3 & 28.3 & 96.0 \\

PruneSID 
& 74.1 & 57.1 & 52.4 & 68.4 & 54.1 & 84.3 & 1366.1 & 59.5 & 54.3 & 26.5 & 94.2 \\

MMTok
& 75.2 & 58.2 & \textbf{53.8} & 68.8 & 55.8 & 85.6 & 1402.3 & 59.4 & 53.9 & 27.5 & 95.7 \\

QCPruner-Early 
& 74.4 & 57.9 & \textbf{53.8} & 67.7 & 55.3 & 85.8 & 1347.8 & 59.7 & 52.6 & 28.0 & 95.0 \\

\textbf{QCPruner} 
& \textbf{76.9} & \textbf{60.8} & 50.6 & 69.5 & \textbf{56.8} & 86.7 & \textbf{1473.2} & \textbf{63.7} & \textbf{56.5} & 29.1 & \textbf{98.2} \\

\rowcolor{tablegray}
\multicolumn{12}{c}{\textit{32 Tokens ($\downarrow$94.4\%)}} \\
FastV 
& 57.1 & 46.8 & 40.7 & 65.8 & 51.5 & 40.5 & 987.2 & 50.5 & 41.5 & 21.9 & 74.5 \\

VisionZip 
& 67.3 & 51.7 & 52.7 & 68.6 & 53.1 & 68.7 & 1243.8 & 56.8 & 50.2 & 26.3 & 88.5 \\

DART 
& 67.9 & 52.9 & 50.3 & 69.1 & 52.0 & 65.3 & 1297.4 & 58.0 & 48.9 & 22.8 & 87.0 \\

DivPrune 
& 71.2 & 54.9 & 53.4 & 68.7 & 52.9 & 81.5 & 1288.0 & 56.8 & 49.1 & 26.8 & 91.4 \\

CDPruner 
& 73.5 & 56.9 & 53.1 & 69.4 & 53.2 & \textbf{87.7} & 1371.5 & 58.8 & 49.5 & 27.2 & 93.9 \\

PruneSID 
& 70.4 & 54.6 & 52.2 & 67.8 & 52.4 & 79.5 & 1340.1 & 55.3 & 49.4 & \textbf{28.0} & 91.1 \\

MMTok
& 73.1 & 56.2 & \textbf{54.5} & 68.8 & 53.5 & 85.9 & 1350.1 & 58.1 & 49.3 & 27.0 & 93.4 \\

QCPruner-Early 
& 71.6 & 55.3 & 53.4 & 68.5 & 53.6 & 81.9 & 1317.6 & 57.5 & 49.8 & 27.4 & 92.3 \\

\textbf{QCPruner} 
& \textbf{75.2} & \textbf{59.6} & 50.1 & \textbf{70.0} & \textbf{54.6} & 86.4 & \textbf{1418.6} & \textbf{62.2} & \textbf{55.6} & 27.4 & \textbf{96.1} \\

\bottomrule
\end{tabular}
\caption{Full benchmark-wise results on LLaVA-1.5-7B.
We report every obtained benchmark-wise result for the compatible baselines under different retained patch-token targets.
Avg. Rel. is computed with respect to the full-token reference.}
\label{tab:app_llava15_full}
\end{table*}

\paragraph{Budget-dependent pattern.}
QCPruner shows particularly large aggregate gains at several tighter token budgets, reaching 2.1 points on LLaVA-NeXT at 160 tokens, 2.0 points on LLaVA-Video at 1024 tokens, and 9.9 points on Qwen2.5-VL at 128 tokens.
Although the gains are often larger under tighter budgets, this trend is not strictly monotonic. On LLaVA-1.5, the margins at 64 and 32 tokens are similar.
The aggregate gains reflect more consistent performance retention across heterogeneous tasks rather than uniform improvements on every individual benchmark.
For video, all methods are compared under matched nominal total token budgets, but FastVID uses dynamic/global allocation whereas QCPruner prunes frame-wise. Thus, the comparison controls the total token budget but not the allocation strategy.

\begin{table*}[t]
\centering
\small
\setlength{\tabcolsep}{1mm}
\begin{tabular}{lccccccccccc}
\toprule
\textbf{Method} 
& \textbf{VQAv2} 
& \textbf{GQA} 
& \textbf{VizWiz} 
& \textbf{SQA-IMG} 
& \textbf{TextVQA} 
& \textbf{POPE} 
& \textbf{MME} 
& \textbf{MMB-EN} 
& \textbf{MMB-CN} 
& \textbf{MMVet} 
& \textbf{Avg. Rel.} \\
\midrule

\rowcolor{tablegray}
\multicolumn{12}{c}{\textit{All 2880 Tokens (100\%)}} \\
LLaVA-NeXT-7B 
& 81.8 & 64.2 & 57.1 & 70.1 & 64.9 & 86.5 & 1519.0 & 67.4 & 60.6 & 40.4 & 100.0 \\

\rowcolor{tablegray}
\multicolumn{12}{c}{\textit{640 Tokens ($\downarrow$77.8\%)}} \\
FastV & 78.4 & 61.9 & 55.7 & 68.8 & 60.1 & 84.2 & 1494.0 & 66.1 & 59.7 & 38.2 & 96.7 \\
SparseVLM & 79.8 & 62.2 & 54.2 & 68.9 & 60.5 & 86.8 & 1493.7 & \textbf{67.8} & \textbf{60.7} & \textbf{39.4} & 97.8 \\
VisionZip & 79.1 & 61.2 & \textbf{57.5} & 67.9 & 60.1 & 86.0 & 1462.1 & 65.8 & 58.4 & 39.2 & 96.9 \\
DART & 79.4 & 63.2 & 55.9 & 69.3 & \textbf{60.6} & 85.8 & 1497.3 & 66.1 & 59.3 & 38.6 & 97.5 \\
DivPrune & 79.7 & 61.9 & 54.9 & 68.9 & 56.1 & 86.2 & 1476.8 & 65.4 & 59.0 & 35.4 & 95.4 \\
CDPruner & 79.9 & 62.6 & 55.5 & 67.8 & 58.6 & 87.2 & 1467.9 & 65.9 & 57.8 & 38.6 & 96.6 \\
PruneSID & 78.6 & 61.7 & 55.3 & 68.1 & 54.7 & 86.1 & 1488.8 & 64.3 & 57.4 & 32.0 & 93.7 \\
MMTok & 79.3 & 62.6 & 55.5 & 68.4 & 58.9 & 87.0 & \textbf{1504.2} & 65.7 & 57.8 & 37.4 & 96.5 \\
QCPruner-Early & 79.7 & 62.0 & 55.0 & 68.8 & 56.0 & 86.5 & 1479.1 & 65.5 & 59.0 & 36.4 & 95.7 \\
\textbf{QCPruner (Ours)} & \textbf{81.0} & \textbf{63.4} & 55.3 & \textbf{69.5} & 59.3 & \textbf{88.0} & 1488.8 & 67.3 & 60.5 & 38.9 & \textbf{98.1} \\

\rowcolor{tablegray}
\multicolumn{12}{c}{\textit{320 Tokens ($\downarrow$88.9\%)}} \\
FastV & 76.6 & 60.0 & 54.2 & 68.5 & 58.6 & 78.4 & 1367.9 & 64.7 & 56.7 & 36.5 & 93.1 \\
SparseVLM & 75.3 & 58.5 & 52.9 & 67.6 & 56.8 & 82.7 & 1414.6 & 63.9 & 56.7 & 33.5 & 92.0 \\
VisionZip & 76.2 & 59.0 & \textbf{56.7} & 67.3 & \textbf{58.9} & 82.1 & 1409.8 & 62.6 & 55.2 & \textbf{39.5} & 94.1 \\
DART & 78.4 & 61.2 & 54.8 & 68.3 & 58.5 & 83.5 & 1425.8 & 64.8 & 56.6 & 38.7 & 95.0 \\
DivPrune & 77.5 & 60.4 & 54.2 & 67.8 & 54.2 & 84.1 & 1451.4 & 64.4 & 57.8 & 33.4 & 93.0 \\
CDPruner & 78.4 & 61.4 & 55.2 & 67.4 & 57.3 & 87.2 & 1442.7 & 64.8 & 55.7 & 36.7 & 94.7 \\
PruneSID & 77.2 & 60.4 & 54.4 & 67.0 & 54.0 & 84.9 & 1464.0 & 62.8 & 56.6 & 34.5 & 92.9 \\
MMTok & 77.7 & 61.1 & 55.3 & 67.5 & 56.8 & 85.9 & \textbf{1481.1} & 64.1 & 56.6 & 35.5 & 94.4 \\
QCPruner-Early & 77.7 & 60.6 & 54.1 & 68.0 & 55.2 & 84.4 & 1423.4 & 64.2 & 57.3 & 33.5 & 93.0 \\
\textbf{QCPruner (Ours)} & \textbf{80.3} & \textbf{63.2} & 54.0 & \textbf{68.8} & 58.4 & \textbf{88.4} & 1447.5 & \textbf{66.1} & \textbf{59.7} & 37.4 & \textbf{96.6} \\

\rowcolor{tablegray}
\multicolumn{12}{c}{\textit{160 Tokens ($\downarrow$94.4\%)}} \\
FastV & 67.8 & 53.5 & 50.7 & 68.2 & 52.6 & 62.3 & 1183.8 & 59.9 & 53.0 & 30.2 & 83.4 \\
VisionZip & 71.4 & 55.3 & \textbf{55.9} & 68.1 & \textbf{56.1} & 74.9 & 1291.7 & 59.6 & 52.8 & 31.6 & 88.0 \\
DART & 73.6 & 57.4 & 52.8 & \textbf{68.5} & 55.2 & 77.3 & 1370.2 & 60.0 & 53.0 & 31.6 & 88.9 \\
DivPrune & 75.0 & 58.8 & 53.5 & 68.2 & 52.6 & 81.5 & 1362.1 & 63.0 & 55.7 & 31.1 & 90.2 \\
CDPruner & 76.7 & 60.7 & 54.7 & 67.0 & 55.5 & 87.9 & 1421.8 & 63.6 & 54.5 & 33.5 & 92.8 \\
PruneSID & 74.1 & 58.1 & 54.5 & 67.2 & 52.2 & 80.1 & 1406.9 & 61.6 & 53.7 & 33.6 & 90.1 \\
MMTok & 75.7 & 60.0 & 55.8 & 67.5 & 54.6 & 83.9 & 1408.0 & 63.4 & 54.6 & 32.9 & 91.9 \\
QCPruner-Early & 75.2 & 59.2 & 53.6 & 67.3 & 53.9 & 82.3 & 1388.6 & 63.1 & 55.7 & 30.9 & 90.6 \\
\textbf{QCPruner (Ours)} & \textbf{79.2} & \textbf{62.3} & 52.5 & \textbf{68.5} & 56.0 & \textbf{88.4} & \textbf{1424.3} & \textbf{65.5} & \textbf{58.8} & \textbf{35.9} & \textbf{94.9} \\
\bottomrule
\end{tabular}
\caption{Full generalization results on LLaVA-NeXT-7B.
We report complete benchmark-wise results under different retained-token budgets.
Best results among reported entries under each budget are in bold.}
\label{tab:app_llava_next_full}
\end{table*}

\begin{table*}[t]
\centering
\small
\setlength{\tabcolsep}{1mm}
\begin{tabular}{llccccccc}
\toprule
\textbf{Method} 
& \textbf{Granularity}
& \textbf{MVBench} 
& \textbf{LongVideoBench} 
& \multicolumn{4}{c}{\textbf{Video-MME}} 
& \textbf{Rel.} \\
\cmidrule(lr){3-3}
\cmidrule(lr){4-4}
\cmidrule(lr){5-8}
& 
& \textbf{test} 
& \textbf{val} 
& \textbf{w/o Sub.} 
& \textbf{Short} 
& \textbf{Medium} 
& \textbf{Long} 
&  \\
\midrule

\rowcolor{tablegray}
\multicolumn{9}{c}{\textit{All $64 \times 169$ Patch Tokens (100\%)}} \\
LLaVA-Video-7B 
& -- 
& 60.8 & 58.9
& 64.3 
& 77.3 & 62.4 & 53.2 
& 100.0 \\

\rowcolor{tablegray}
\multicolumn{9}{c}{\textit{Target: 4096 Patch Tokens, equivalent to $64 \times 64$ ($\downarrow$62.1\%)}} \\
FastV        
& frame-wise
& 59.4 & 57.8 
& 63.4 
& 74.8 & 62.9 & 52.6 
& 98.1 \\
DivPrune     
& frame-wise
& 57.8 & 58.5
& 62.7 
& 74.7 & 61.7 & 51.7 
& 97.3 \\
FastVID
& dynamic/global
& \textbf{60.5} & 58.2
& 64.3 
& \textbf{76.4} & 62.0 & \textbf{54.6} 
& 99.4 \\
QCPruner (Ours) 
& frame-wise
& 60.1 & \textbf{58.6} 
& \textbf{64.5} 
& 76.1 & \textbf{63.0} & 54.2 
& \textbf{99.6} \\

\rowcolor{tablegray}
\multicolumn{9}{c}{\textit{Target: 2048 Patch Tokens, equivalent to $64 \times 32$ ($\downarrow$81.1\%)}} \\
FastV        
& frame-wise
& 56.8 & 55.3 
& 62.2 
& 73.1 & 61.8 & 51.8 
& 94.7 \\
DivPrune     
& frame-wise
& 56.3 & 57.0 
& 60.4 
& 71.6 & 58.3 & 51.3 
& 94.4 \\
FastVID
& dynamic/global
& \textbf{59.4} & 56.3 
& 62.9 
& 74.4 & \textbf{61.9} & 52.3 
& 97.0 \\
QCPruner (Ours) 
& frame-wise
& 59.2 & \textbf{58.0} 
& \textbf{63.3} 
& \textbf{74.7} & \textbf{61.9} & \textbf{53.2} 
& \textbf{98.1} \\

\rowcolor{tablegray}
\multicolumn{9}{c}{\textit{Target: 1024 Patch Tokens, equivalent to $64 \times 16$ ($\downarrow$90.5\%)}} \\
FastV        
& frame-wise
& 54.8 & 52.8 
& 59.0 
& 69.3 & 57.8 & 49.9 
& 90.5 \\
DivPrune     
& frame-wise
& 55.0 & 53.3
& 59.3 
& 69.6 & 58.2 & 50.2
& 91.1 \\
FastVID
& dynamic/global
& 58.1 & 55.3 
& 60.1 
& 70.9 & 58.7 & 50.8 
& 94.3 \\
QCPruner (Ours) 
& frame-wise
& \textbf{58.5} & \textbf{57.0} 
& \textbf{61.6} 
& \textbf{72.7} & \textbf{60.3} & \textbf{51.7}
& \textbf{96.3} \\

\bottomrule
\end{tabular}
\caption{Complete results on LLaVA-Video-7B with 64 frames per video, evaluated on MVBench, LongVideoBench, and Video-MME.
All methods are matched by nominal total patch-token budget. Non-frame-wise methods such as FastVID are matched by total rather than per-frame token counts.
Best reported results within each budget are bolded.
Rel. is the equal-weight average of relative scores on MVBench test, LongVideoBench val, and Video-MME without subtitles. 
The Short/Medium/Long columns are diagnostic breakdowns and are excluded from Rel.}
\label{tab:app_video_full}
\end{table*}

\begin{table*}[t]
\centering
\small
\setlength{\tabcolsep}{1mm}
\begin{tabular}{lcccccccc}
\toprule
\textbf{Method} 
& \textbf{TextVQA} 
& \textbf{ChartQA} 
& \textbf{AI2D} 
& \textbf{OCRBench} 
& \textbf{MME} 
& \textbf{MMB-EN} 
& \textbf{MMB-CN} 
& \textbf{Avg. Rel.} \\
\midrule

\rowcolor{tablegray}
\multicolumn{9}{c}{\textit{All 1296 Tokens (100\%)}} \\
Qwen2.5-VL-7B 
& 82.6 & 82.1 & 84.1 & 738 
& 2332.8 & 83.5 & 80.7 
& 100.0 \\

\rowcolor{tablegray}
\multicolumn{9}{c}{\textit{512 Tokens ($\downarrow$60.5\%)}} \\
FastV        
& \textbf{82.4} & 79.3 & 82.5 & 692 
& 2336.4 & 82.4 & 78.8 
& 97.8 \\
DivPrune     
& 80.0 & 72.8 & 82.1 & 678 
& 2325.5 & 81.9 & 78.4 
& 95.7 \\
VisionZip     
& 80.4 & \textbf{81.1} & \textbf{83.0} & 698
& \textbf{2356.9} & \textbf{83.6} & 79.5 
& 98.4 \\
MMTok
& 78.2 & 75.8 & 82.6 & 667
& 2274.5 & 82.5 & 79.1
& 95.7 \\
\textbf{QCPruner (Ours)} 
& 81.6 & 80.5 & \textbf{83.0} & \textbf{709} 
& 2319.1 & 83.2 & \textbf{79.6} 
& \textbf{98.5} \\

\rowcolor{tablegray}
\multicolumn{9}{c}{\textit{256 Tokens ($\downarrow$80.2\%)}} \\
FastV        
& 79.9 & 70.8 & 78.8 & 579 
& 2299.8 & 80.6 & 76.8 
& 92.2 \\
DivPrune     
& 74.2 & 62.7 & 80.5 & 544 
& 2247.3 & 81.4 & 77.6 
& 89.4 \\
VisionZip     
& 73.9 & 72.1 & 80.8 & 586 
& \textbf{2339.8} & 81.5 & 78.3 
& 92.5 \\
MMTok
& 72.5 & 68.2 & 81.5 & 581
& 2268.4 & 71.0 & 77.5
& 89.3 \\
\textbf{QCPruner (Ours)} 
& \textbf{80.7} & \textbf{77.2} & \textbf{82.1} & \textbf{672} 
& 2336.1 & \textbf{82.1} & \textbf{79.0} 
& \textbf{96.7} \\

\rowcolor{tablegray}
\multicolumn{9}{c}{\textit{128 Tokens ($\downarrow$90.1\%)}} \\
FastV        
& 74.0 & 54.2 & 71.4 & 420 
& 2110.3 & 73.7 & 70.1 
& 80.4 \\
DivPrune     
& 65.4 & 48.4 & 76.0 & 411 
& 2125.0 & 76.8 & 74.8 
& 80.0 \\
VisionZip     
& 62.8 & 56.6 & 77.4 & 457 
& 2194.2 & 78.4 & 75.4 
& 82.9 \\
MMTok
& 61.7 & 49.8 & 78.1 & 435
& 2177.9 & 78.3 & 75.1
& 81.1 \\
\textbf{QCPruner (Ours)} 
& \textbf{77.3} & \textbf{70.7} & \textbf{80.4} & \textbf{583} 
& \textbf{2336.9} & \textbf{81.6} & \textbf{78.6} 
& \textbf{92.8} \\

\bottomrule
\end{tabular}
\caption{Full Qwen2.5-VL-7B generalization results.
Avg. Rel. denotes the average relative performance retention with respect to the full-token Qwen2.5-VL-7B reference.
Best results among reported entries under each budget are in bold.}
\label{tab:app_qwen_full}
\end{table*}

\section{Ablation Studies}
\label{app:additional_ablation}

This section reports query-anchor construction overhead, utility-floor calibration, bilateral target--representative weighting, keyword-count sensitivity, controlled MMR/DPP selector comparisons, and pruning-layer stability under separate calibration.

\subsection{Query-Anchor Construction Overhead}
\label{app:keyword_efficiency}

\paragraph{Query-anchor construction.}
The measured query-anchor construction time is \textbf{1.67 ms/sample} on LLaVA-NeXT-7B, including \textbf{1.65 ms/sample} for keyword extraction and span matching and \textbf{0.02 ms/sample} for anchor attachment.
The one-time keyword-extractor loading cost is \textbf{26.89 ms} and is excluded from the per-sample latency.
These measurements characterize post-initialization anchor-construction overhead and exclude cue computation, dense affinity computation, and greedy selection.

\subsection{Independent Selection of the Utility Floor}
\label{app:rho_ablation}

Rather than tuning $\rho$ on any benchmark in the reported evaluation suite, we select it once for each model family or setting using separate calibration data and then freeze it across downstream datasets and token budgets.
We sweep $\rho\in\{0,0.1,\ldots,1.0\}$, with $\rho=0$ corresponding to disabling the utility floor.
The image-model sweeps use the full MMStar dataset~\cite{mmstar} without subsampling.
For video, we use a fixed 600-question subset of the NExT-QA~\cite{nextqa} \texttt{nextqa\_mc\_test} split, obtained with random seed 42 by stratified sampling of 200 questions from each of its three question categories.
Each sweep uses the middle reported token budget: 64 tokens for LLaVA-1.5, 256 for Qwen2.5-VL, and 2048 for LLaVA-Video.
The value selected on LLaVA-1.5 is transferred to LLaVA-NeXT without further tuning.

We use MMStar for image calibration because it provides broad multimodal coverage while remaining disjoint from the reported downstream test suite.
We use NExT-QA for video calibration because its three question categories support stratified sampling while remaining disjoint from the reported video benchmarks.
For both calibration datasets, the sweep score is multiple-choice accuracy averaged over the corresponding calibration set.

For any two tokens $i$ and $j$ with $0\leq\rho<1$,
\begin{equation}
\hat{u}_i-\hat{u}_j=(1-\rho)(u_i-u_j),
\label{eq:app_rho_ordering}
\end{equation}
so the transformation preserves pairwise utility orderings while contracting score gaps by a factor of $1-\rho$.
At $\rho=1$, all calibrated utilities equal one and the selector reduces to coverage-only selection. We include this value only as a diagnostic endpoint.

\begin{table}[t]
\centering
\small
\setlength{\tabcolsep}{1.4mm}
\begin{tabular}{cccc}
\toprule
$\boldsymbol{\rho}$
& \textbf{LLaVA-1.5}
& \textbf{Qwen2.5-VL}
& \textbf{LLaVA-Video} \\
& \textbf{MMStar}
& \textbf{MMStar}
& \textbf{NExT-QA} \\
\midrule
0.0 & \textbf{\underline{0.3294}} & 0.5756 & 0.3783 \\
0.1 & 0.3245 & 0.5885 & 0.3883 \\
0.2 & 0.3268 & 0.5925 & 0.3850 \\
0.3 & 0.3273 & 0.5901 & 0.3617 \\
0.4 & 0.3293 & \textbf{\underline{0.6018}} & 0.3800 \\
0.5 & 0.3159 & 0.6012 & \textbf{\underline{0.3933}} \\
0.6 & 0.3225 & 0.5887 & 0.3783 \\
0.7 & 0.3179 & 0.5958 & 0.3683 \\
0.8 & 0.3249 & 0.6005 & 0.3667 \\
0.9 & 0.3226 & 0.5894 & 0.3333 \\
1.0 & 0.3213 & 0.5902 & 0.3750 \\
\bottomrule
\end{tabular}
\caption{Utility-floor calibration on separate calibration benchmarks that are not included as reported test sets. Bold indicates the deployed coefficient, and underline indicates the highest observed calibration score in each column.}
\label{tab:app_rho_ablation}
\end{table}

Table~\ref{tab:app_rho_ablation} reports the complete calibration sweep.
On LLaVA-1.5, disabling the utility floor with $\rho=0$ yields the highest observed calibration score of 0.3294, narrowly above 0.3293 at $\rho=0.4$.
Because no positive value improves upon the no-floor setting, we retain $\rho=0$ for the LLaVA image family.
Qwen2.5-VL reaches its highest observed calibration score of 0.6018 at $\rho=0.4$, while LLaVA-Video reaches 0.3933 at $\rho=0.5$.
We therefore use $\rho=0$, $0.4$, and $0.5$ for the LLaVA image, Qwen2.5-VL, and LLaVA-Video settings, respectively, and keep these values fixed across the reported downstream benchmarks and token budgets.

\subsection{Target- and Representative-Side Utility Weighting}
\label{app:bilateral_utility}

We isolate the two utility factors on LLaVA-1.5-7B at 64/576 patch tokens and Qwen2.5-VL-7B at 128/1296 and 256/1296 patch tokens.
For an edge from target token $i$ to candidate representative $j$, we compare visual-only affinity $c_{i,j}$, target-side weighting $c_{i,j}\hat u_i$, representative-side weighting $c_{i,j}\hat u_j$, and the full bilateral form $c_{i,j}\hat u_i\hat u_j$.
Within each backbone--budget setting, the query anchors, fused utility, post-layer-7 pruning position, candidate set, greedy coverage selector, and token budget are fixed.
The deployed utility floor is $\rho=0$ for LLaVA and $\rho=0.4$ for all query-weighted Qwen settings. Visual-only rows use no utility weighting.

\begin{table}[t]
\centering
\small
\setlength{\tabcolsep}{0.35mm}
\textit{(a) LLaVA-1.5-7B, 64/576 tokens}\par
\begin{tabular}{lrrrrrrrr}
\toprule
\textbf{Edge Weight} & \textbf{GQA} & \textbf{POPE} & \textbf{MME} & \textbf{SQA} & \textbf{Text} & \textbf{M-E} & \textbf{M-C} & \textbf{Rel.} \\
\midrule
$c_{i,j}$
& 60.2 & 82.7 & 1425.4 & 68.2 & 54.0 & 62.3 & 54.8 & 95.7 \\
$c_{i,j}\hat u_i$
& 60.6 & 86.1 & 1448.9 & 68.5 & 55.2 & 62.7 & 56.1 & 97.3 \\
$c_{i,j}\hat u_j$
& 60.3 & \textbf{86.7} & 1449.2 & \textbf{69.6} & 56.2 & 63.3 & 56.3 & 98.0 \\
$c_{i,j}\hat u_i\hat u_j$ (full)
& \textbf{60.8} & \textbf{86.7} & \textbf{1473.2} & 69.5 & \textbf{56.8} & \textbf{63.7} & \textbf{56.5} & \textbf{98.6} \\
\bottomrule
\end{tabular}

\par
\medskip
\textit{(b) Qwen2.5-VL-7B}\par
\begin{tabular}{lrrrrrrrr}
\toprule
\textbf{Edge Weight} & \textbf{Text} & \textbf{Chart} & \textbf{AI2D} & \textbf{OCR} & \textbf{MME} & \textbf{M-E} & \textbf{M-C} & \textbf{Rel.} \\
\midrule
\multicolumn{9}{c}{\textit{256/1296 tokens}} \\
$c_{i,j}$
& 77.6 & 70.2 & 81.2 & 640 & 2286.0 & 81.4 & 77.3 & 93.4 \\
$c_{i,j}\hat u_i$
& 78.8 & 70.9 & 81.1 & 652 & 2293.3 & \textbf{82.3} & 78.1 & 94.3 \\
$c_{i,j}\hat u_j$
& 80.4 & \textbf{77.2} & \textbf{82.4} & \textbf{676} & 2327.6 & \textbf{82.3} & 78.9 & \textbf{96.7} \\
$c_{i,j}\hat u_i\hat u_j$ (full)
& \textbf{80.7} & \textbf{77.2} & 82.1 & 672 & \textbf{2336.1} & 82.1 & \textbf{79.0} & \textbf{96.7} \\
\midrule
\multicolumn{9}{c}{\textit{128/1296 tokens}} \\
$c_{i,j}$
& 70.5 & 60.0 & 78.8 & 526 & 2254.4 & 79.6 & 76.9 & 87.2 \\
$c_{i,j}\hat u_i$
& 71.8 & 61.2 & 79.4 & 529 & 2242.2 & 80.0 & 77.3 & 87.9 \\
$c_{i,j}\hat u_j$
& 77.2 & 70.1 & \textbf{80.7} & \textbf{589} & 2325.7 & 81.0 & 78.2 & 92.6 \\
$c_{i,j}\hat u_i\hat u_j$ (full)
& \textbf{77.3} & \textbf{70.7} & 80.4 & 583 & \textbf{2336.9} & \textbf{81.6} & \textbf{78.6} & \textbf{92.8} \\
\bottomrule
\end{tabular}
\caption{Utility-factor ablations across three backbone--budget settings. Text/Chart/OCR and M-E/M-C denote TextVQA/ChartQA/OCRBench and MMBench-EN/CN, respectively. Rel. denotes average relative performance retention against the corresponding full-token reference. Bold indicates the best displayed value within each backbone--budget block.}
\label{tab:app_bilateral_utility}
\end{table}

Table~\ref{tab:app_bilateral_utility} reports the task-wise utility-factor ablations.
On LLaVA, target-side and representative-side weighting improve Rel. from 95.7\% to 97.3\% and 98.0\%, respectively, while the bilateral form further reaches 98.6\% and improves five of seven displayed task scores over representative-only weighting.
On Qwen at 256 tokens, the corresponding Rel. values are 93.4\%, 94.3\%, 96.7\%, and 96.7\%. The bilateral and representative-only forms each lead on three tasks and tie on one.
At 128 tokens, the values are 87.2\%, 87.9\%, 92.6\%, and 92.8\%, while the bilateral form leads representative-only on five of seven tasks.
Overall, representative-side utility provides the stronger one-sided contribution across all three settings.
Bilateral weighting yields an additional aggregate gain on LLaVA and Qwen-128, while its incremental benefit over representative-only weighting remains modest and setting-dependent.

\subsection{Sensitivity to the Number of Query Keywords}
\label{app:anchor_hyper}

On LLaVA-1.5-7B at 64/576 patch tokens, we vary only the maximum keyword count while fixing the extractor, span matching, post-layer-7 pruning position, equal-weight cue fusion, and coverage selector.

\begin{table}[!t]
\centering
\small
\setlength{\tabcolsep}{0.35mm}
\begin{tabular}{@{}crrrrrrrr@{}}
\toprule
\textbf{Max. Keywords} & \textbf{GQA} & \textbf{POPE} & \textbf{MME} & \textbf{Text} & \textbf{SQA} & \textbf{M-E} & \textbf{M-C} & \textbf{Rel.} \\
\midrule
2 & 60.6 & 86.8 & 1446.1 & 55.9 & 69.1 & 63.6 & 55.7 & 97.8 \\
4 & 60.8 & 86.7 & 1480.9 & 56.6 & 69.2 & 63.5 & 56.8 & 98.6 \\
6 & 60.8 & 86.7 & 1473.2 & 56.8 & 69.5 & 63.7 & 56.5 & 98.6 \\
8 & 60.8 & 86.7 & 1474.7 & 57.0 & 69.5 & 63.5 & 56.4 & 98.6 \\
\bottomrule
\end{tabular}
\caption{Sensitivity to the maximum number of extracted query keywords on LLaVA-1.5-7B at 64/576 patch tokens. Text/M-E/M-C abbreviate TextVQA/MMBench-EN/MMBench-CN. Rel. is the seven-task average relative performance against the unpruned model.}
\label{tab:app_anchor_hyper}
\end{table}

Table~\ref{tab:app_anchor_hyper} shows that Avg. Rel. increases from 97.8\% with two keywords to 98.6\% with four keywords and remains at 98.6\% at the reported precision with six or eight.
Thus, performance is stable once at least four keywords are available, and the default maximum of six does not depend on a narrow optimum.

\subsection{Controlled Comparison of Subset Selectors}
\label{app:dpp_transplant}

\begin{table*}[t]
\centering
\small
\setlength{\tabcolsep}{1.1mm}
\textit{(a) Qwen2.5-VL-7B}\par
\begin{tabular}{lcrrrrrrrr}
\toprule
\textbf{Selector} & \textbf{Tokens} & \textbf{Text} & \textbf{Chart} & \textbf{AI2D} & \textbf{OCR} & \textbf{MME} & \textbf{M-E} & \textbf{M-C} & \textbf{Rel.} \\
\midrule
Full & 1296 & 82.6 & 82.1 & 84.1 & 738 & 2332.8 & 83.5 & 80.7 & 100.0 \\
\midrule
Top-$K$ & 128 & 49.1 & 25.1 & 75.4 & 185 & 2217.9 & 79.3 & 74.4 & 69.6 \\
MMR & 128 & 75.2 & 62.2 & 79.2 & 513 & 2315.6 & 80.2 & 76.8 & 88.7 \\
CDPruner-DPP & 128 & 77.0 & 62.8 & 79.4 & 533 & 2286.2 & 79.5 & 75.3 & 89.0 \\
MMTok-style & 128 & 76.0 & 62.4 & 80.1 & 566 & 2286.2 & 79.3 & 76.5 & 89.7 \\
Visual-only Cov. & 128 & 70.5 & 60.0 & 78.8 & 526 & 2254.4 & 79.6 & 76.9 & 87.2 \\
QCPruner & 128 & \textbf{77.3} & \textbf{70.7} & \textbf{80.4} & \textbf{583} & \textbf{2336.9} & \textbf{81.6} & \textbf{78.6} & \textbf{92.8} \\
\midrule
Top-$K$ & 256 & 63.5 & 38.7 & 79.3 & 310 & 2260.9 & \textbf{82.2} & 76.5 & 78.6 \\
MMR & 256 & 78.8 & 71.5 & 81.5 & 599 & 2306.3 & 82.1 & 78.0 & 93.5 \\
CDPruner-DPP & 256 & 79.7 & 71.2 & 81.4 & 615 & 2318.4 & 81.1 & 77.4 & 93.7 \\
MMTok-style & 256 & 80.4 & 72.6 & 82.0 & \textbf{677} & 2272.3 & 80.8 & 77.7 & 95.1 \\
Visual-only Cov. & 256 & 77.6 & 70.2 & 81.2 & 640 & 2286.0 & 81.4 & 77.3 & 93.4 \\
QCPruner & 256 & \textbf{80.7} & \textbf{77.2} & \textbf{82.1} & 672 & \textbf{2336.1} & 82.1 & \textbf{79.0} & \textbf{96.7} \\
\bottomrule
\end{tabular}
\par
\textit{(b) LLaVA-1.5-7B}\par
\begin{tabular}{lcrrrrrrrr}
\toprule
\textbf{Selector} & \textbf{Tokens} & \textbf{GQA} & \textbf{POPE} & \textbf{Text} & \textbf{M-E} & \textbf{M-C} & \textbf{MME} & \textbf{SQA} & \textbf{Rel.} \\
\midrule
Full & 576 & 61.9 & 85.9 & 58.2 & 64.7 & 58.1 & 1506.5 & 69.5 & 100.0 \\
\midrule
Top-$K$ & 64 & 60.8 & 86.2 & 53.5 & 62.2 & 56.4 & 1398.9 & 68.9 & 96.5 \\
MMR & 64 & 60.8 & 86.6 & 56.3 & 63.1 & \textbf{57.0} & 1459.5 & \textbf{69.6} & 98.3 \\
CDPruner-DPP & 64 & \textbf{61.2} & 86.6 & \textbf{56.9} & 63.4 & 56.8 & 1467.3 & 69.3 & \textbf{98.6} \\
MMTok-style & 64 & 60.5 & 85.8 & 56.7 & 63.5 & 56.0 & 1434.2 & 69.4 & 97.8 \\
Visual-only Cov. & 64 & 60.2 & 82.7 & 54.0 & 62.3 & 54.8 & 1425.4 & 68.2 & 95.7 \\
QCPruner & 64 & 60.8 & \textbf{86.7} & 56.8 & \textbf{63.7} & 56.5 & \textbf{1473.2} & 69.5 & \textbf{98.6} \\
\bottomrule
\end{tabular}
\caption{Controlled selector comparison across Qwen2.5-VL-7B and LLaVA-1.5-7B. Text/Chart/OCR and M-E/M-C denote TextVQA/ChartQA/OCRBench and MMBench-EN/CN, respectively. Rel. denotes seven-task average relative performance retention with respect to the corresponding full-token reference. Candidate patches, query anchors, fused cues, pruning position, utility floor, and token budget are fixed within each backbone--budget setting. Only the subset selector changes. Bold indicates the best selector result within each budget. Visual-only Cov. is an internal control, while MMTok-style is an objective-level backend transplant and is distinct from the complete MMTok system.}
\label{tab:app_qwen_selector_tasks}
\end{table*}

\paragraph{Visual-token MMR.}
For $R$ candidates, let $K'=\min(K,R)$ and normalize QCPruner's fused query utility as
\[
\bar r_i=
\frac{r_i-r_{\min}}
{\max(r_{\max}-r_{\min},\varepsilon)},
\qquad \varepsilon=10^{-6}.
\]
We apply the same backbone-level utility floor used by the corresponding QCPruner setting,
\[
q_i^{\mathrm{MMR}}
=
\rho_b+(1-\rho_b)\bar r_i,
\]
with $\rho_b=0.4$ for Qwen2.5-VL and $\rho_b=0$ for LLaVA-1.5.
The visual-similarity matrix is the same candidate-wise affinity matrix used by QCPruner.
Let $\widetilde C=[\operatorname{float32}(C)]_+$ elementwise and $d_i^{(0)}=0$.
The controlled MMR selector is
\begin{equation}
\begin{aligned}
j_t
&=\arg\max_{i\notin\mathcal K_{t-1}}
\left[
0.5q_i^{\mathrm{MMR}}
-
0.5d_i^{(t-1)}
\right],\\
d_i^{(t)}
&=
\max\{d_i^{(t-1)},\widetilde C_{i,j_t}\},
\qquad t=1,\ldots,K'.
\end{aligned}
\label{eq:app_mmr}
\end{equation}
Thus, MMR uses the same query utility, visual affinity, candidate set, and token budget as QCPruner, while replacing only the subset-selection rule.

\paragraph{MMTok-style objective transplant.}
For the Qwen2.5-VL and LLaVA-1.5 selector comparisons, we additionally replace the subset objective with MMTok's additive text--vision and vision--vision coverage formulation~\cite{mmtok}, while retaining the same candidate patches, post-layer-7 pruning position, query anchors, upstream representations, and token budgets.
The objective form and all MMTok-specific hyperparameters follow the official configuration without modification or task-specific retuning.
Because this transplant operates on QCPruner's in-decoder inputs rather than MMTok's original pre-decoder pipeline, we denote it \emph{MMTok-style} and distinguish it from the complete MMTok system in the benchmark tables.

\paragraph{CDPruner-DPP transplant.}
For the controlled selector comparison, we transplant CDPruner's conditional-DPP/Fast-MAP backend~\cite{cdpruner} while fixing the candidate patches, post-layer-7 pruning position, keyword-matched anchors, fused query utility, and token budget.
We use $K\in\{128,256\}$ for Qwen2.5-VL and $K=64$ for LLaVA-1.5.
This \emph{CDPruner-DPP} setting is a selector-level transplant and is distinct from the complete CDPruner system reported in the benchmark tables.
Its DPP/Fast-MAP selection logic follows the official CDPruner implementation without algorithmic modification.\footnote{\url{https://github.com/Theia-4869/CDPruner}}

Let $S\in\mathbb{R}^{R\times R}$ denote the candidate-wise visual-similarity matrix and $u_i$ the fused query utility.
The transplanted quality mapping and DPP kernel are
\begin{equation}
\begin{aligned}
\bar u_i
&=
\frac{u_i-u_{\min}}
{\max(u_{\max}-u_{\min},\varepsilon)},\\
q_i
&=
\rho+(1-\rho)\bar u_i,\\
L^{\mathrm{DPP}}
&=
\operatorname{diag}(\mathbf q)
S
\operatorname{diag}(\mathbf q),
\end{aligned}
\label{eq:app_dpp_kernel}
\end{equation}
where $\rho=\rho_b$ and $\varepsilon=10^{-6}$.
Thus, QCPruner utility replaces CDPruner's native relevance signal, while the DPP similarity treatment, determinant objective, and Fast-MAP subset-selection backend remain unchanged.

Table~\ref{tab:app_qwen_selector_tasks} reports the controlled selector results.
On Qwen2.5-VL, QCPruner achieves 92.8 and 96.7 Rel. at 128 and 256 tokens, exceeding CDPruner-DPP by 3.8 and 3.0 points and MMTok-style by 3.1 and 1.6 points, respectively.
QCPruner leads all seven displayed tasks at 128 tokens and five at 256 tokens. MMTok-style and Top-$K$ lead OCRBench and MMBench-EN, respectively, at the latter budget.
On LLaVA-1.5 at 64 tokens, CDPruner-DPP and QCPruner both reach 98.6 Rel. at the reported precision, compared with 97.8 for MMTok-style.
The task-level leaders are distributed across QCPruner, CDPruner-DPP, and MMR on three, two, and two tasks, respectively.
Compared with visual-only coverage, query-weighted coverage improves Rel. by 5.6, 3.3, and 2.9 points at Qwen-128, Qwen-256, and LLaVA-64, respectively.
Because the visual affinity, coverage selector, and token budget are fixed in these comparisons, these gains isolate the contribution of query weighting within the coverage formulation.
Overall, QCPruner achieves higher average retention than the controlled MMTok-style backend in all three settings and than the DPP backend on Qwen, while QCPruner and DPP perform similarly on LLaVA.

\subsection{Pruning-Layer Stability on Separate Calibration Data}
\label{app:layer_stability}

Before evaluating the reported downstream test benchmarks, we analyze pruning-layer stability on all 1,500 MMStar samples using LLaVA-1.5-7B at 64/576 patch tokens.
We use unpruned forward passes to maintain a fixed visual-token population across layers, while keeping the query anchors, equal-weight cue fusion, and scoring procedure fixed with $\rho=0$.
All layer indices use zero-based decoder indexing. Thus, the deployed pruning position at $\ell=7$ is after decoder layer 7 and before layer 8.

For sample $n$, let $\mathbf u_n^\ell$ denote the fused utility vector at layer $\ell$.
We define a sample-specific late-layer consensus proxy as
\begin{equation}
\bar{\mathbf u}^{\mathrm{late}}_n
=
\frac{1}{3}\sum_{r=9}^{11}\mathbf u_n^r.
\label{eq:app_late_utility_reference}
\end{equation}
Layer-wise rank stability is measured by the average within-sample Spearman correlation,
\begin{equation}
S_\ell
=
\frac{1}{1500}\sum_{n=1}^{1500}
\operatorname{Spearman}
\!\left(
\mathbf u_n^\ell,
\bar{\mathbf u}^{\mathrm{late}}_n
\right).
\label{eq:app_layer_spearman}
\end{equation}
Here, $\operatorname{Spearman}(\mathbf a,\mathbf b)$ is the
Pearson correlation between their rank-transformed entries, so
it measures agreement in visual-token ordering rather than raw
score scale, with larger values indicating more stable rankings.
We compute this correlation within each sample and then average
it across the 1,500 samples.

To measure agreement among the highest-utility tokens, let
$\mathcal T_n^\ell=\operatorname{Top64}(\mathbf u_n^\ell)$ and
$\mathcal T_n^{\mathrm{late}}=\operatorname{Top64}(\bar{\mathbf u}^{\mathrm{late}}_n)$.
We compute the average per-sample Jaccard overlap
\begin{equation}
J_\ell
=
\frac{1}{1500}\sum_{n=1}^{1500}
\frac{
|\mathcal T_n^\ell\cap\mathcal T_n^{\mathrm{late}}|
}{
|\mathcal T_n^\ell\cup\mathcal T_n^{\mathrm{late}}|
}.
\label{eq:app_layer_jaccard}
\end{equation}
Cue-wise correlations are computed analogously using the corresponding cue average over layers 9--11 as the consensus proxy.
Figure~2 in the main paper visualizes the full trajectories, and Table~\ref{tab:app_layer_stability_full} reports the corresponding numerical values.

\begin{table}[!htbp]
\centering
\small
\setlength{\tabcolsep}{0.6mm}
\begin{tabular}{crrrr}
\toprule
\textbf{Layer} & \textbf{Fused} & \textbf{Top-64} & \textbf{Align.} & \textbf{Hidden} \\
\textbf{$\ell$} & \textbf{Spear.} & \textbf{Jac.} & \textbf{Spear.} & \textbf{Spear.} \\
\midrule
1  & 0.565 & 0.247 & 0.724 & 0.293 \\
2  & 0.593 & 0.279 & 0.778 & 0.352 \\
3  & 0.627 & 0.322 & 0.787 & 0.407 \\
4  & 0.672 & 0.383 & 0.823 & 0.454 \\
5  & 0.690 & 0.418 & 0.855 & 0.439 \\
6  & 0.730 & 0.470 & 0.880 & 0.476 \\
\textbf{7} & \textbf{0.863} & \textbf{0.567} & \textbf{0.909} & \textbf{0.688} \\
8  & 0.906 & 0.636 & 0.942 & 0.809 \\
\rowcolor{tablegray}
9  & 0.962 & 0.764 & 0.963 & 0.929 \\
\rowcolor{tablegray}
10 & 0.985 & 0.849 & 0.985 & 0.971 \\
\rowcolor{tablegray}
11 & 0.968 & 0.788 & 0.961 & 0.948 \\
\bottomrule
\end{tabular}
\caption{Mean layer-wise agreement over 1,500 MMStar samples. Fused and cue-specific Spearman correlations use the corresponding per-sample layer-9--11 consensus proxies, and Top-64 reports Jaccard overlap. Bold marks the deployed pruning position at layer 7 rather than the maximum value. Gray rows form the consensus reference and are not independent evaluation points. All layer indices are zero-based.}
\label{tab:app_layer_stability_full}
\end{table}

The fused Spearman correlation increases gradually through layer 6 and then rises from 0.730 to 0.863 at layer 7 (+0.133), followed by a smaller increase to 0.906 at layer 8 (+0.043).
Top-64 Jaccard shows a similar transition, increasing from 0.470 to 0.567 between layers 6 and 7 (+0.097) and to 0.636 at layer 8.
Later layers reach higher agreement, as expected given that layers 9--11 define the late-layer consensus proxy.

The cue decomposition further clarifies this transition.
Alignment-cue stability is already high by layer 5 and increases smoothly from 0.855 to 0.942 across layers 5--8.
In contrast, hidden-state similarity rises sharply from 0.476 at layer 6 to 0.688 at layer 7 (+0.212).
This stronger change in the semantic cue coincides with the increase in fused-ranking stability at layer 7.

We therefore use layer 7 as an early practical pruning point where fused-ranking stability has increased substantially.
Pruning after decoder layer 7 retains eight completed decoder layers for cue formation while allowing all subsequent layers to process the reduced sequence.
Although layer 8 shows higher agreement, it requires one additional full-token decoder layer before pruning.
Layer 7 therefore provides a practical balance between cue maturity and the remaining opportunity for computation reduction.
We fix this pruning position before downstream evaluation and use it across the evaluated backbone--budget settings without test-set retuning.

\subsection{Cue and Query-Anchor Ablation at 64 Tokens}
\label{app:cue_anchor_budget}

The main paper reports the cue and query-anchor ablation under the most aggressive 32/576-token setting.
Here, we repeat the comparison at 64/576 patch tokens while fixing the post-layer-7 hook, population-coverage selector, query-anchor construction, and all other settings.

\begin{table}[t]
\centering
\small
\setlength{\tabcolsep}{0.35mm}
\begin{tabular}{lrrrrrrrr}
\toprule
\textbf{Configuration} & \textbf{SQA} & \textbf{GQA} & \textbf{POPE} & \textbf{MME} & \textbf{Text} & \textbf{M-E} & \textbf{M-C} & \textbf{Rel.} \\
\midrule
Sim.+Anc. & 69.1 & 60.1 & \textbf{86.7} & 1463.7 & 55.2 & 62.6 & 55.4 & 97.4 \\
QK+Anc. & 69.5 & 60.4 & 85.4 & \textbf{1478.1} & 56.3 & \textbf{63.7} & 56.1 & 98.1 \\
Fusion+Prompt & \textbf{69.6} & 60.1 & 86.3 & 1470.7 & 56.6 & 63.4 & \textbf{56.6} & 98.3 \\
Fusion+Anc. & 69.5 & \textbf{60.8} & \textbf{86.7} & 1473.2 & \textbf{56.8} & \textbf{63.7} & 56.5 & \textbf{98.6} \\
\bottomrule
\end{tabular}
\caption{Cue and query-anchor ablations on LLaVA-1.5-7B at 64/576 patch tokens. Anc. denotes keyword-matched anchors. Text/M-E/M-C abbreviate TextVQA/MMBench-EN/MMBench-CN. Rel. is the seven-task average relative performance against the unpruned model. Bold marks the best pruned result.}
\label{tab:app_cue_anchor_budget}
\end{table}

Table~\ref{tab:app_cue_anchor_budget} shows that the fused-anchor variant reaches 98.6 Avg. Rel., compared with 98.1 for QK alignment alone and 97.4 for hidden-state similarity alone.
Thus, equal-weight fusion provides a modest aggregate gain at this budget, while the task-level leaders remain distributed across variants.
Replacing matched anchors with the full prompt reduces Avg. Rel. from 98.6 to 98.3.

\section{Transferability of Query Anchors}
\label{app:transferability}

To examine whether keyword-matched query anchors can transfer beyond QCPruner, we integrate them into SparseVLM, an existing text-guided visual token pruning method.
Specifically, we replace SparseVLM's original textual condition with our keyword-matched anchors while keeping the remaining pruning pipeline unchanged.
This controlled replacement isolates the contribution of textual conditioning within the SparseVLM pipeline.

\begin{table}[!b]
\centering
\small
\setlength{\tabcolsep}{0.6mm}
\begin{tabular}{@{}lrrrrrrrr@{}}
\toprule
\textbf{Method} 
& \textbf{GQA} 
& \textbf{POPE} 
& \textbf{SQA} 
& \textbf{Text}
& \textbf{MME} 
& \textbf{M-E}
& \textbf{M-C}
& \textbf{Rel.} \\
\midrule
SparseVLM 
& 53.7 & 77.4 & 69.6 & 53.4
& 1290.1 & 59.3 & 52.4 & 90.9 \\
$+$ anchors
& \textbf{54.0} & \textbf{78.6} & \textbf{69.7} & \textbf{53.5}
& \textbf{1330.0} & \textbf{60.3} & \textbf{52.8} & \textbf{91.9} \\
$\Delta$ 
& +0.3 & +1.2 & +0.1 & +0.1
& +39.9 & +1.0 & +0.4 & +1.0 \\
\bottomrule
\end{tabular}
\caption{Transfer of keyword-matched query anchors to SparseVLM on LLaVA-1.5-7B at 64/576 patch tokens. Avg. Rel. is computed over the seven displayed benchmarks.}
\label{tab:app_transferability}
\end{table}

As shown in Table~\ref{tab:app_transferability}, replacing SparseVLM's original textual condition with keyword-matched anchors improves all seven benchmarks and increases Avg. Rel. from 90.9\% to 91.9\%.
Overall, these results show that the proposed query anchors are not tied to QCPruner's coverage selector and can transfer to another text-guided pruning pipeline.

\section{Limitations and Broader Impact}
\label{app:limitations}
\label{app:broader_impact}

\paragraph{Limitations.}
QCPruner relies on keyword-matched anchors produced by an external extractor with English stop-word filtering.
Short, highly referential, multilingual, or OCR-oriented prompts may therefore provide incomplete textual conditions, although the deterministic fallback reduces complete matching failures.
The population-coverage selector also requires dense pairwise affinity computation and direct $O(KN_v^2)$ greedy selection, which introduces additional pruning overhead.
Moreover, the shared pruning position is selected using LLaVA-1.5 on MMStar and then transferred across evaluated backbones, while the utility floor remains calibrated at the model-family level rather than fixed globally.
Performance gains are also task- and setting-dependent, with occasional regressions on individual benchmarks.
Finally, downstream QA retention does not directly measure region-level grounding quality, which remains an important direction for further evaluation.

\paragraph{Broader Impact.}
This work aims to improve the practicality of multimodal large language models by retaining compact visual-token subsets that preserve query-relevant information.
Reducing the visual-token sequence processed by later decoder layers may lower model-side computation while maintaining strong task performance.
However, pruning can alter model behavior when relevant visual evidence is removed.
Applications requiring high reliability, including medical, legal, and other safety-critical settings, should therefore validate pruning behavior carefully before deployment.

\section{Qualitative Motivation Examples}
\label{app:qualitative_examples}

These examples complement the quantitative analyses by visualizing layer-wise query-conditioned responses, query-anchor conditioning, and the complementary behavior of the two utility cues. Each heat map is independently normalized and is intended for qualitative interpretation.

\begin{table*}[!t]
    \centering
    \small
    \setlength{\tabcolsep}{0.35mm}
    \begin{tabular}{@{}ccccccccc@{}}
    \textbf{Layer 2} & \textbf{Layer 3} & \textbf{Layer 4} & \textbf{Layer 5} & \textbf{Layer 6} & \textbf{Layer 7} & \textbf{Layer 8} & \textbf{Layer 9} & \textbf{Layer 10} \\
    \multicolumn{9}{c}{\textbf{(a) ``book'' example (sample 130)}} \\
    \includegraphics[width=0.102\textwidth,height=0.088\textheight,keepaspectratio]{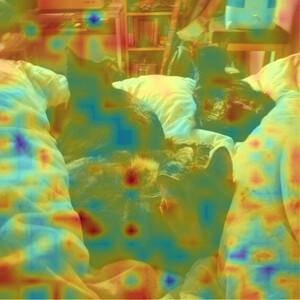} &
    \includegraphics[width=0.102\textwidth,height=0.088\textheight,keepaspectratio]{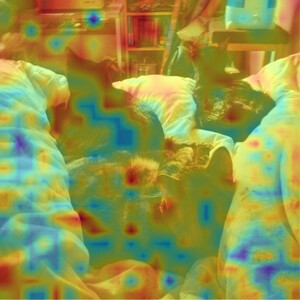} &
    \includegraphics[width=0.102\textwidth,height=0.088\textheight,keepaspectratio]{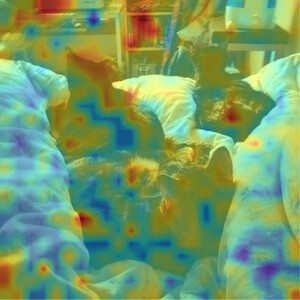} &
    \includegraphics[width=0.102\textwidth,height=0.088\textheight,keepaspectratio]{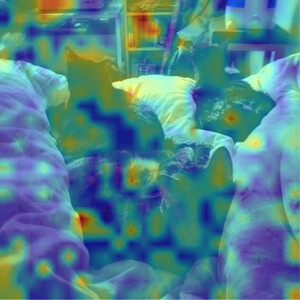} &
    \includegraphics[width=0.102\textwidth,height=0.088\textheight,keepaspectratio]{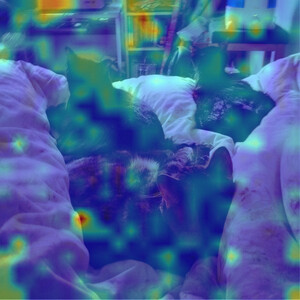} &
    \includegraphics[width=0.102\textwidth,height=0.088\textheight,keepaspectratio]{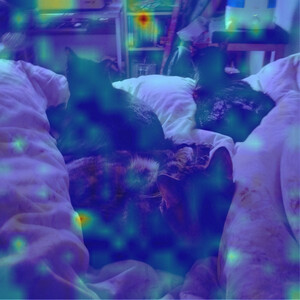} &
    \includegraphics[width=0.102\textwidth,height=0.088\textheight,keepaspectratio]{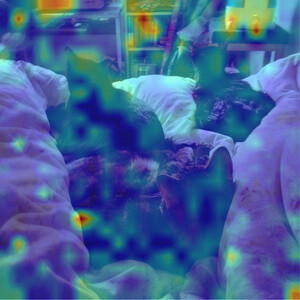} &
    \includegraphics[width=0.102\textwidth,height=0.088\textheight,keepaspectratio]{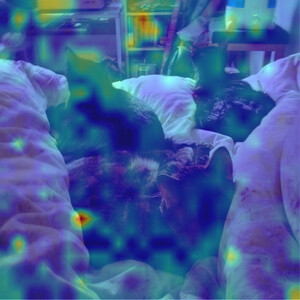} &
    \includegraphics[width=0.102\textwidth,height=0.088\textheight,keepaspectratio]{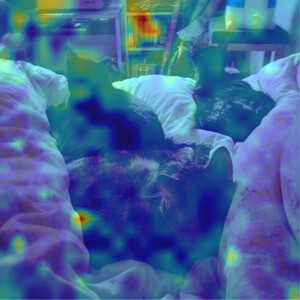} \\
    \multicolumn{9}{c}{\textbf{(b) ``mouse'' example (sample 134)}} \\
    \includegraphics[width=0.102\textwidth,height=0.088\textheight,keepaspectratio]{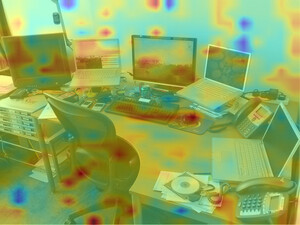} &
    \includegraphics[width=0.102\textwidth,height=0.088\textheight,keepaspectratio]{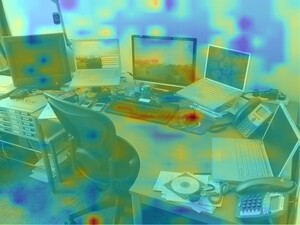} &
    \includegraphics[width=0.102\textwidth,height=0.088\textheight,keepaspectratio]{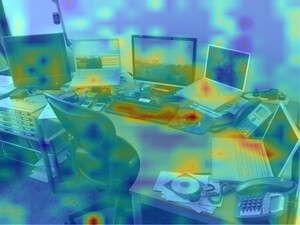} &
    \includegraphics[width=0.102\textwidth,height=0.088\textheight,keepaspectratio]{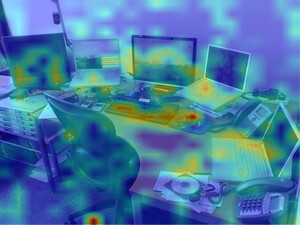} &
    \includegraphics[width=0.102\textwidth,height=0.088\textheight,keepaspectratio]{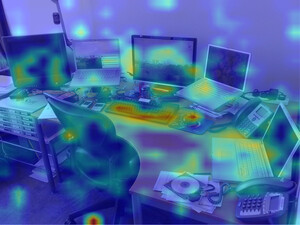} &
    \includegraphics[width=0.102\textwidth,height=0.088\textheight,keepaspectratio]{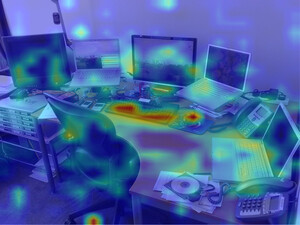} &
    \includegraphics[width=0.102\textwidth,height=0.088\textheight,keepaspectratio]{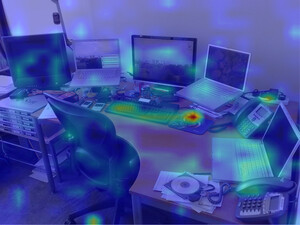} &
    \includegraphics[width=0.102\textwidth,height=0.088\textheight,keepaspectratio]{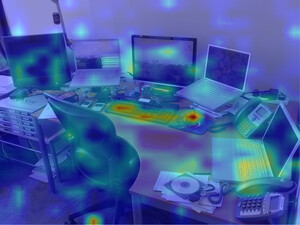} &
    \includegraphics[width=0.102\textwidth,height=0.088\textheight,keepaspectratio]{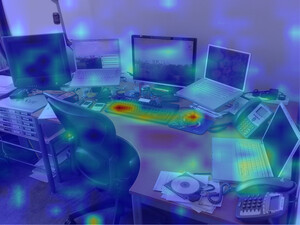} \\
    \multicolumn{9}{c}{\textbf{(c) ``sink'' example (sample 138)}} \\
    \includegraphics[width=0.102\textwidth,height=0.088\textheight,keepaspectratio]{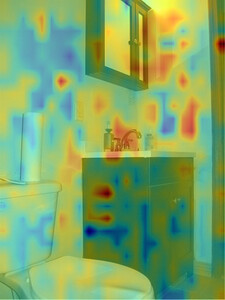} &
    \includegraphics[width=0.102\textwidth,height=0.088\textheight,keepaspectratio]{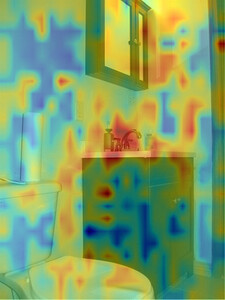} &
    \includegraphics[width=0.102\textwidth,height=0.088\textheight,keepaspectratio]{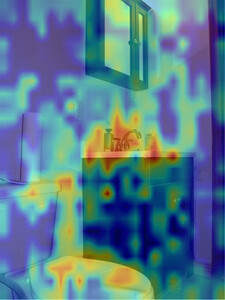} &
    \includegraphics[width=0.102\textwidth,height=0.088\textheight,keepaspectratio]{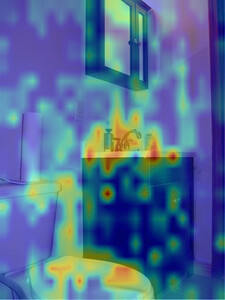} &
    \includegraphics[width=0.102\textwidth,height=0.088\textheight,keepaspectratio]{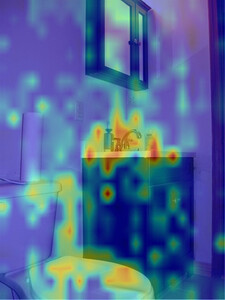} &
    \includegraphics[width=0.102\textwidth,height=0.088\textheight,keepaspectratio]{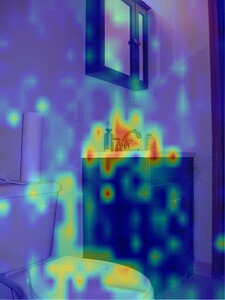} &
    \includegraphics[width=0.102\textwidth,height=0.088\textheight,keepaspectratio]{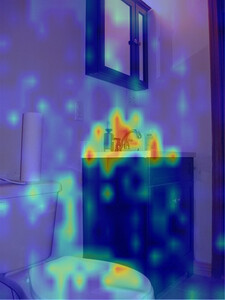} &
    \includegraphics[width=0.102\textwidth,height=0.088\textheight,keepaspectratio]{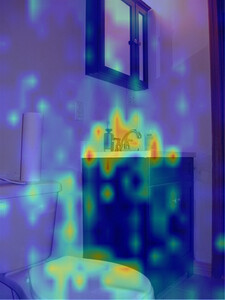} &
    \includegraphics[width=0.102\textwidth,height=0.088\textheight,keepaspectratio]{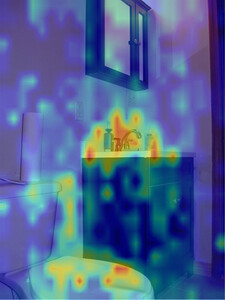}
    \end{tabular}

    \captionof{figure}{Layer-wise fused responses for the ``book,'' ``mouse,'' and ``sink'' examples. Each heading denotes the response after the named decoder layer, with layers 2--10 running from left to right. Red and yellow denote larger displayed responses within each independently normalized map. These examples illustrate the layer transition rather than provide a grounding metric.}
    \label{fig:app_layer_sequences}
\end{table*}

\begin{table*}[!t]
    \centering
    \includegraphics[width=0.94\textwidth,keepaspectratio]{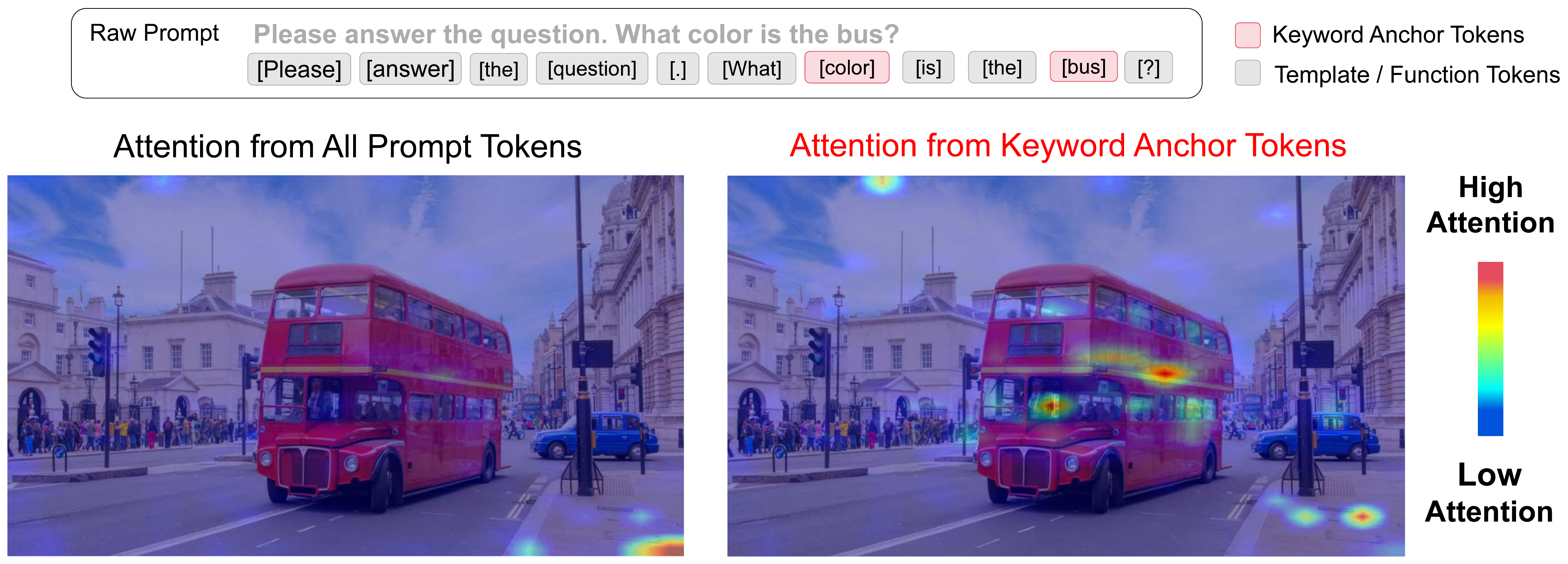}
    \captionof{figure}{Query-anchor motivation. Full-prompt alignment is compared with alignment conditioned on matched content phrases. Each heat map is independently normalized for visualization.}
    \label{fig:app_anchor_motivation}
\end{table*}

\begin{table*}[!t]
    \centering
    \includegraphics[width=0.94\textwidth,keepaspectratio]{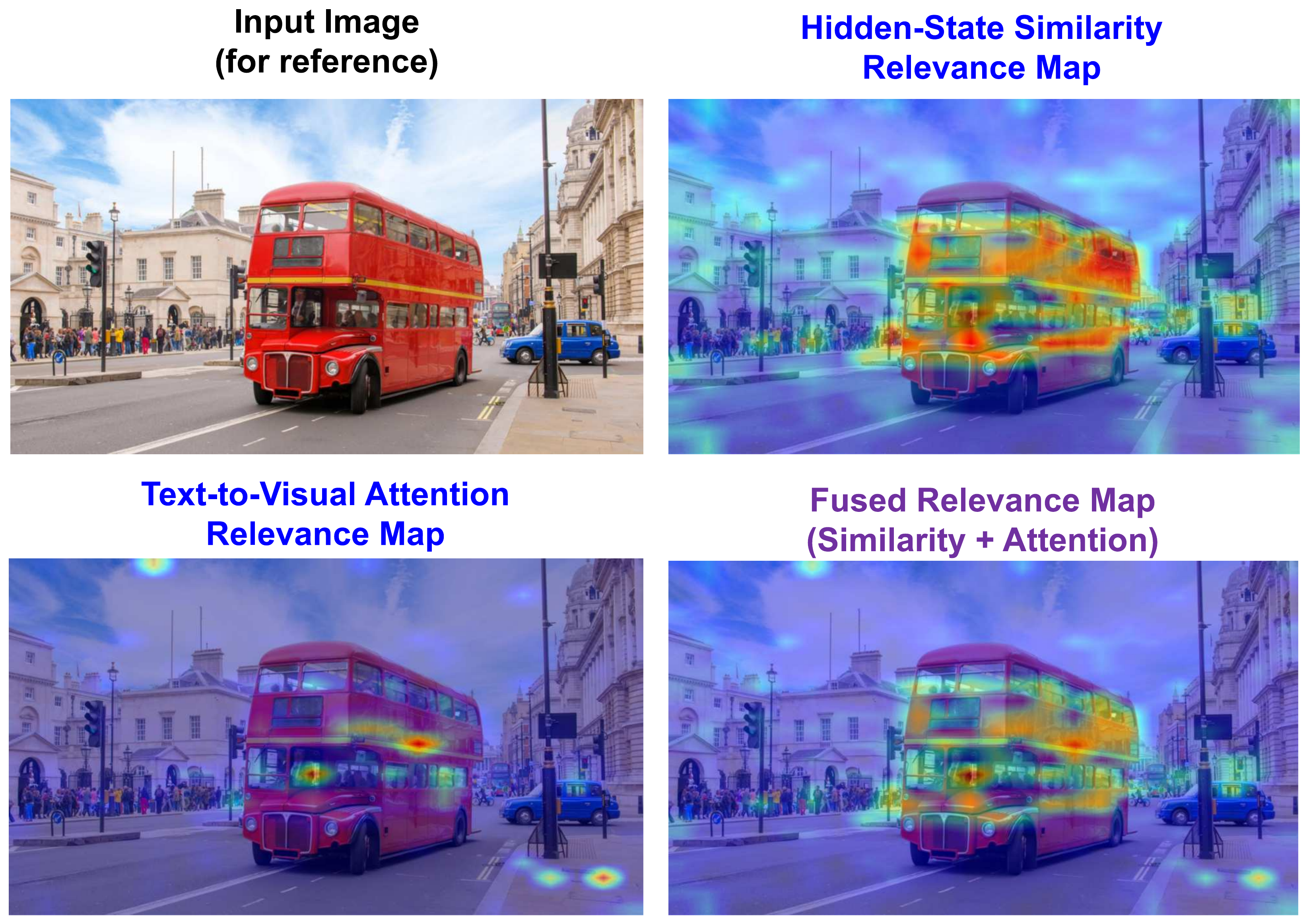}
    \captionof{figure}{Complementary query-conditioned cues. Hidden-state cosine similarity and head-averaged pre-softmax query--key alignment exhibit distinct response patterns, which QCPruner combines after normalization using equal weights.}    \label{fig:app_cue_motivation}
\end{table*}

\subsection{Qualitative View of the Layer Transition}
\label{app:layer_selection}

Figure~\ref{fig:app_layer_sequences} visualizes the equal-weight fused query-conditioned response across successive decoder layers for three examples.
Early-layer responses are comparatively diffuse, while later layers show more concentrated response patterns.
This qualitative evolution is consistent with the increase in ranking stability observed around layer 7 in Appendix~\ref{app:layer_stability}.

\subsection{Query Anchors}

Prompts mix task-relevant content with system scaffolding, punctuation, and generic instruction words.
As illustrated in Figure~\ref{fig:app_anchor_motivation}, full-prompt alignment can be spatially dispersed, whereas matched content phrases such as ``color'' and ``bus'' produce more focused query-conditioned responses.
These visualizations motivate using keyword-matched phrases as focused textual anchors while retaining the fallback mechanism described in Appendix~\ref{app:implementation}.

\subsection{Complementary Query-Conditioned Cues}

Figure~\ref{fig:app_cue_motivation} contrasts hidden-state cosine similarity, head-averaged pre-softmax query--key alignment, and their equal-weight fusion for the same query anchors.
The two cues exhibit complementary response patterns: similarity tends to distribute responses across semantically related regions, whereas alignment can be more concentrated around a smaller set of locations.
QCPruner therefore combines their normalized responses with equal weights, consistent with the quantitative cue ablation reported in the main paper.
\end{document}